\documentclass[letterpaper, 10 pt, conference]{ieeeconf}  % Comment this line out if you need a4paper

\IEEEoverridecommandlockouts                              % This command is only needed if 
\usepackage{amssymb}
\usepackage{amsmath}
\usepackage[english]{babel}
\usepackage{float}
\usepackage{graphicx}
\usepackage{hyperref}
\usepackage{mathtools}

\usepackage{filecontents}
\usepackage[noadjust]{cite}
\usepackage[font=footnotesize,labelfont=footnotesize]{subcaption}
\usepackage[font=footnotesize,labelfont=footnotesize]{caption}
\usepackage{subcaption}
\usepackage{caption}
\usepackage{comment}
\usepackage{authblk}
\usepackage{multirow}
\usepackage{environ}
\usepackage{graphicx}

\usepackage{amsthm, algorithm, algorithmicx} %amsthm, 
\usepackage[noend]{algpseudocode}
\usepackage[english]{babel}

\let\labelindent\relax

\usepackage[inline]{enumitem}

\newtheorem{innercustomgeneric}{\customgenericname}
\providecommand{\customgenericname}{}
\newcommand{\newcustomtheorem}[2]{%
  \newenvironment{#1}[1]
  {%
   \renewcommand\customgenericname{#2}%
   \renewcommand\theinnercustomgeneric{##1}%
   \innercustomgeneric
  }
  {\endinnercustomgeneric}
}

\newcustomtheorem{customthm}{Theorem}
\newcustomtheorem{customlemma}{Lemma}
\newcustomtheorem{customprop}{Proposition}

\theoremstyle{plain}  % to use together with amsthm package
\newtheorem{theorem}{Theorem}
\newtheorem{lemma}[theorem]{Lemma}
\newtheorem{definition}[theorem]{Definition}
\newtheorem{remark}{Remark}
\newtheorem{proposition}[theorem]{Proposition}

\algnewcommand\algorithmicinput{\textbf{Input:}}
\algnewcommand\algorithmicoutput{\textbf{Output:}}
\algnewcommand\Input{\item[\algorithmicinput]}%
\algnewcommand\Output{\item[\algorithmicoutput]}%

\DeclareMathOperator*{\argmin}{arg\,min}
\DeclareMathOperator*{\argmax}{arg\,max}

\usepackage{xcolor}

\newcommand{\remove}[1]{}

\title{\LARGE \bf
Minimally Constrained Multi-Robot Coordination with Line-of-sight \\ Connectivity Maintenance
}

\author{Yupeng Yang$^1$, Yiwei Lyu$^2$, and Wenhao Luo$^1$% <-this % stops a space
\thanks{$^*$This work was supported in part by the Faculty Research Grant award at the University of North Carolina at Charlotte.}
\thanks{$^1$The authors are with the Department of Computer Science, University of North Carolina at Charlotte, Charlotte, NC 28223, USA. Email: {\tt \{yyang52, wenhao.luo\}@uncc.edu}}
\thanks{$^{2}$The author is with the Department of Electrical and Computer Engineering, Carnegie Mellon University, Pittsburgh, PA 15213, USA. Email: {\tt yiweilyu@andrew.cmu.edu}}%
}

\begin{document}

\maketitle
\thispagestyle{empty}
\pagestyle{empty}

%%%%%%%%%%%%%%%%%%%%%%%%%%%%%%%%%%%%%%%%%%%%%%%%%%%%%%%%%%%%%%%%%%%%%%%%%%%%%%%%
\begin{abstract}
In this paper, we consider a team of mobile robots executing simultaneously multiple behaviors by different subgroups, while maintaining global and subgroup line-of-sight (LOS) network connectivity that minimally constrains the original multi-robot behaviors. The LOS connectivity between pairwise robots is preserved when two robots stay within the limited communication range and their LOS remains occlusion-free from static obstacles while moving. By using control barrier functions (CBF) and minimum volume enclosing ellipsoids (MVEE), we first introduce the LOS connectivity barrier certificate (LOS-CBC) to characterize the state-dependent admissible control space for pairwise robots, from which their resulting motion will keep the two robots LOS connected over time. We then propose the Minimum Line-of-Sight Connectivity Constraint Spanning Tree (MLCCST) as a step-wise bilevel optimization framework to jointly optimize (a) the minimum set of LOS edges to actively maintain, and (b) the control revision with respect to a nominal multi-robot controller due to LOS connectivity maintenance. As proved in the theoretical analysis, this allows the robots to improvise the optimal composition of LOS-CBC control constraints that are least constraining around the nominal controllers, and at the same time enforce the global and subgroup LOS connectivity through the resulting preserved set of pairwise LOS edges. The framework thus leads to robots staying as close to their nominal behaviors, while exhibiting dynamically changing LOS-connected network topology that provides the greatest flexibility for the existing multi-robot tasks in real time. We demonstrate the effectiveness of our approach through simulations with up to 64 robots.
\end{abstract}
%%%%%%%%%%%%%%%%%%%%%%%%%%%%%%%%%%%%%%%%%%%%%
% In this paper, we consider a team of mobile robots executing simultaneously multiple behaviors by different subgroups, while maintaining global and subgroup line-of-sight (LOS) network connectivity that minimally constrains the original multi-robot behaviors. 
% The LOS connectivity between pairwise robots is preserved when two robots are within the limited communication range and their LOS is occlusion-free from the static obstacles over time.
% By using control barrier functions (CBF) and minimum volume enclosing ellipsoids (MVEE), we introduce the LOS connectivity barrier certificate characterizing admissible control space for pairwise robots to enforce LOS connectivity over time. 
% \yy{With that, we present a step-wise bilevel multi-robot optimization framework that (1) selects the least constraining LOS connectivity subgraph to maintain at each time step, and (2) computes the control revision with respect to a nominal multi-robot controller subject to safety and incurred LOS connectivity constraints by (1).
% With our proposed Minimum Line-of-Sight Connectivity Constraint Spanning Tree (MLCCST) approach and the theoretical analysis, the solution to the framework is proved to enable the resulting multi-robot behaviors minimally deviated from the nominal ones, while satisfying the required global and subgroup LOS connectivity constraints.}
% We demonstrate the effectiveness of our approach through simulations with up to 64 robots.
%%%%%%%%%%%%%%%%%%%%%%%%%%%%%%%%%%
\section{Introduction}
% In many multi-robot applications, 
Connectivity maintenance is critical to ensure effective information exchange among robots. It is often achieved through constraining robots' motion due to their limited communication range, so that the proximity-based communication graph remains as one connected component when robots are moving, commonly referred to as maintaining \emph{global connectivity}~\cite{sabattini2013decentralized,yang2010decentralized,williams2015global, luo2019voronoi, ong2021network,capelli2020connectivity}.  
% \yy{To ensure effective information exchange among robots under limited communication capability, 
% connectivity maintenance concerns about maintaining the proximity-based communication graph as one connected component (commonly referred as maintaining \emph{global connectivity}~\cite{sabattini2013decentralized,yang2010decentralized,williams2015global,ong2021network,capelli2020connectivity}) by constraining the collective robot motion. }
Most of the existing methods use either local methods or global methods. In local methods, connectivity is maintained by preserving the initial connectivity graph topology over time~\cite{zavlanos2007flocking,dimarogonas2008decentralized,ji2007distributed}, while in global methods the algebraic connectivity of the communication graph is maintained by a secondary connectivity controller that keeps the second smallest eigenvalue of the graph Laplacian positive at all times~\cite{sabattini2013decentralized,yang2010decentralized,williams2015global,tateo2018multiagent, capelli2020connectivity}. 

However, these works assume a simplistic connectivity model where pairwise robots can always communicate as long as they are within the limited communication range, and may not apply to some realistic environments.
% In scenarios such as search and rescue, 
For instance, occlusions from solid obstacles in between robots may cause disruptions in information exchange, e.g. thick walls may cutoff the Bluetooth communication.
% , and further lead the multi-agent systems to unsafe states with potential collisions. 
Therefore, it is desired to take realistic factors such as the ‘Line-of-sight’ (LOS) problem into account~\cite{sun2020optimal,tuck2021dec, shetty2021decentralized}.
% that addresses visibility-based connectivity maintenance.
In \cite{sun2020optimal},
it is formulated as a nonlinear constraint in a mixed integer problem to preserve the LOS connectivity,
% which enforces the minimum distance from an obstacle to the line segment between two connected agents always greater than the radius of the obstacle.
while it is not applicable for real-time computation and cannot scale up well with increased number of agents. In \cite{tuck2021dec}, explicit visibility subgraphs and blind-spots in the workspace are computed to allow agents to navigate safely within each others' sights, but when new agents previously not in the same visibility graph emerge in LOS, an emergency brake will be performed while recomputing visibility subgraphs, leading to overly conservative maneuver.
LOS-aware formation and connectivity control are presented in \cite{gao2019velocity, shetty2021decentralized} where robot behaviors are dominated by the connectivity-oriented design. As a result, robots may not progress toward achieving the primary goal when tasked to spread out and disperse over a wide area.
% comes at a cost of lower task efficiency. 
Thus how to maintain LOS connectivity while providing the greatest flexibility for robots 
% to execute the 
% in a minimally invasive manner with respect to 
% task-related nominal controllers 
is an important challenge.
% worth investigating.
% \wl{a motivating example of los constraints may be needed (reviewer 10). }

% To realize collision avoidance and connectivity maintenance minimally constraining the task-related behaviors for MAS, Control Barrier Function (CBF) based methods \cite{ames2019control} 
% %like Barrier Certificate 
% with provable theoretical guarantees have been widely used that allow a more admissible space for the constrained multi-robot motion~\cite{li2018formally,pierpaoli2020sequential,wang2016multi}. 
% The network connectivity is enforced through a pre-defined composition of communication edges to retain.
% However, these approaches can not handle possible communication cutoff due to occlusions from obstacles. 
% To the best of our knowledge, there is no prior work addressing the LOS problem using methods like Barrier Certificate for minimally invasive MAS control.
% Another gap identified in current works addressing MAS connectivity maintenance is that, 
On the other hand, 
addressing complicated communication topology requirements beyond global connectivity is also a critical challenge.
In many multi-robot applications, it usually requires robots to perform more than one task with different designated subgroups by simultaneously executing multiple behaviors~\cite{lin2021online,luo2020behavior, luo2019minimum, luo2020minimally} or sequences of behaviors~\cite{nagavalli2017automated, li2018formally,pierpaoli2020sequential} for a set of sub-tasks.
For example, consider a team of autonomous ground and aerial vehicles split into multiple subgroups based on their capabilities and tasked to simultaneously explore widely separated regions with various local tasks executed by different subgroups. Achieving efficient local collaboration requires robots in the same subgroup to be connected as one component and global connectivity across different subgroups is also needed for global situation awareness.
How to ensure LOS connectivity both within each subgroup and across subgroups remains also as a challenge.

% existing methods are often not able to handle complicated communication topology requirements such as ensuring both global and subgroup connectivity, where agents split in different subgroups are required to pursue distinct tasks simultaneously, calling for  .
%  \cite{luo2020behavior} presents a Minimum Connectivity Constraint Spanning Tree algorithm to preserve both subgroup and global connectivity in a minimally invasive manner to the nominal controllers.
% In real-world, MAS deployment usually requires robots to perform more than one task with different designated subgroups by simultaneously executing multiple behaviors~\cite{lin2021online,luo2020behavior} and sequences of behaviors~\cite{li2018formally,pierpaoli2020sequential} for a set of sub-tasks, namely behavior mixing. To this end, how to ensure LOS connectivity within each subgroup, across subgroups and within MAS remains as a challenge. 

In this work, we propose a novel method to consider collision avoidance and global and subgroup LOS connectivity maintenance in multi-robot coordination. 
% \wl{the overall goal}
The \textbf{key contribution} is three-fold. \textbf{\textit{First}}, we propose a novel notion of Line-of-sight Connectivity Barrier Certificate using Control Barrier Functions (CBF) \cite{ames2019control}  
% the LOS maintenance problem is addressed by the novel composition of Line-of-sight Connectivity Barrier Certificate, 
that defines admissible control space for preserving pairwise LOS connectivity with formal guarantees.
% which provides a formally provable guarantee for LOS connectivity over time. 
\textbf{\textit{Second}}, by integrating LOS Connectivity Barrier Certificate and a graph theoretic approach, we formulate the minimally constrained LOS-aware coordination problem as a step-wise bi-level optimization problem, and propose
% by combining LOS Connectivity Barrier Certificate, Safety Barrier Certificate \cite{wang2017safety}, and a graph theoretic approach, the problem is formulated as a step-wise bi-level optimization and 
the Minimum Line of Sight Connectivity Constraint Spanning Tree (MLCCST) method to \emph{co-optimize} (a) the global and subgroup connectivity topology to enforce, and (b) control deviation from nominal controllers subject to the LOS connectivity constraints, thus providing greatest flexibility for the nominal task-related robot motion.  
% In this way, the task efficiency is maximized subject to the computed minimally constraining LOS connectivity requirements. 
\textbf{\textit{Third}}, we supply formal theoretical analysis 
% (also with more detailed discussions in appendix \cite{yupeng}) 
on our proposed MLCCST approach.
% , and the solution to the framework is theoretically proved with claimed advantages. 
% Performance comparison between our proposed method and two baseline methods are demonstrated on up to 64 robots in simulation to show its effectiveness and strong scalability.

\section{Preliminaries}\label{section2}
Consider a robotic team $\mathcal{S}$ consisting of $N$ mobile robots. Each robot $i\in \{1,\ldots,N\}$ is located at the position $\mathbf{x}_i \in \mathbb{R}^d$ with dynamics 
% $\dot{\mathbf{x}}_i$ affine in control as
$\dot{\mathbf{x}}_i =f_i(\mathbf{x}_i)+g_i(\mathbf{x}_i)\mathbf{u}_i$,
where $f_i: \mathbb{R}^{d} \rightarrow \mathbb{R}^{d}$ and $g_i: \mathbb{R}^{d} \rightarrow \mathbb{R}^{d\times q}$ are locally Lipschitz continuous
and $\mathbf{u}_i\in\mathbb{R}^q$. The workspace consists of free space and occupied space $\mathcal{C}_{\mathrm{obs}}=\bigcup\limits_{k=1}^K \mathcal{O}_k$ by $K$ static polyhedral obstacles $\mathcal{O}_k\subset \mathbb{R}^d,\forall k$.\\
\noindent
\textbf{Communication models}: 
% Each robot is able to connect and communicate directly with other robots within its spatial proximity. 
Denote $\mathcal{G} = (\mathcal{V},\mathcal{E})$ 
as the \textbf{\textit{communication graph}} of the robotic team, where each node $v \in \mathcal{V}$ represents a robot. Conventionally, if $||\mathbf{x}_i-\mathbf{x}_j||\leq R_\mathrm{c}$ with $R_\mathrm{c}\in\mathbb{R}$ as the limited communication range, then it is assumed the two robots are \textbf{\emph{connected}} and can communicate with the undirected edge $(v_i,v_j) \in \mathcal{E}$ (i.e.$(v_i,v_j) \in \mathcal{E} \Longleftrightarrow (v_j,v_i) \in \mathcal{E}$). However, in obstacle-populated environments, the robots can only effectively communicate with others that are \textit{not only} within the limited communication \textit{but also} in line-of-sight (LOS) free from occlusions by all the obstacles, i.e. two robots $v_i,v_j$ are \textbf{\textit{LOS connected}} with the undirected LOS edge $(v_i,v_j)\in \mathcal{E}^\mathrm{los}$ if $  \mathbf{x}_i(1-\beta)+\mathbf{x}_j\beta \notin \mathcal{C}_{\mathrm{obs}},\;\forall \beta\in [0,1], (v_i,v_j) \in \mathcal{E}$ (Occlusion-free condition). Hence, we define the {\textit{\textbf{LOS communication graph}}} as $\mathcal{G}^\mathrm{los} = (\mathcal{V},\mathcal{E}^\mathrm{los})\subseteq \mathcal{G}$ with $\mathcal{E}^\mathrm{los}\subseteq \mathcal{E}$ as the set of LOS communication edges. In this case, maintaining LOS connectivity between two robots in $(v_i,v_j)\in \mathcal{E}^\mathrm{los}$ requires satisfying (a) connectivity constraint $||\mathbf{x}_i-\mathbf{x}_j||\leq R_\mathrm{c}$ and (b) occlusion-free condition.
\subsection{Safety and Connectivity Constraints using CBFs}
% It is shown in the previous section that a Control Barrier Certificate can be constructed with a desired set $\mathcal{H}$, which is characterized by a continuously differentiable function $h$. 
% Before addressing occlusion-free LOS condition, 
% We first review how to ensure  safety and connectivity constraints over time using control barrier functions \cite{wang2016multi,ames2019control}. 
Assume the $K$ static polyhedral obstacles can be commonly represented by $F$ discretized obstacles located along the boundary of the static obstacles \cite{hubbard1996approximating,thirugnanam2021duality}. Each discretized obstacle is denoted as $o \in \{1,...,F\}$.
Consider joint robot states $\mathbf{x} = \{\mathbf{x}_1,...,\mathbf{x}_N\} 
\in \mathcal{X} 
\subset \mathbb{R}^{dN}$, the discretized joint obstacle  states $\mathbf{x}^\mathrm{obs}=\{\mathbf{x}^\mathrm{obs}_1,\ldots,\mathbf{x}^\mathrm{obs}_F\}\in\mathcal{X}^\mathrm{obs}\subset \mathbb{R}^{dF}$, the minimum inter-robot safe distance $R_\mathrm{s}\in\mathbb{R}$, the minimum obstacle-robot safe distance $R_\mathrm{obs}\in\mathbb{R}$ and the limited communication range $R_\mathrm{c}$. 
The desired set on $\mathbf{x}$ for any pairwise robots $i$, $j$ and obstacle $o$ satisfying inter-robot or robot-obstacle collision avoidance can be defined as:
% \vspace{-0.2cm}
% \begin{footnotesize}
\begin{align}
h^\mathrm{s}_{i,j}(\mathbf{x}) &= ||\mathbf{x}_i -\mathbf{x}_j||^2 -R_\mathrm{s}^2, \forall i>j, 
\nonumber
\\
%\qquad \qquad
\mathcal{H}^\mathrm{s}_{i,j} &= \{\mathbf{x} \in \mathbb{R}^{dN}: h^\mathrm{s}_{i,j}(\mathbf{x}) \geq 0 \} \label{eq:h_safe}\\
h^\mathrm{obs}_{i,o}(\mathbf{x},\mathbf{x}^\mathrm{obs}) &= ||\mathbf{x}_i -\mathbf{x}^\mathrm{obs}_o||^2 -R_\mathrm{obs}^2, \forall i,o,
\nonumber
\\
%\qquad \qquad
\mathcal{H}^\mathrm{obs}_{i,o} &= \{\mathbf{x} \in \mathbb{R}^{dN}: h^\mathrm{obs}_{i,o}(\mathbf{x},\mathbf{x}^\mathrm{obs}_{o}) \geq 0 \} \label{eq:h_o_safe}
\end{align}
% \end{footnotesize}
And the desired set on $\mathbf{x}$ for any pairwise robots $i$ and $j$ satisfying connectivity constraint can be defined as:
% \vspace{-0.4cm}
% \begin{footnotesize}
\begin{align}
h^\mathrm{c}_{i,j}(\mathbf{x}) &= R_\mathrm{c}^2 - ||\mathbf{x}_i - \mathbf{x}_j||^2, \;\forall (v_i,v_j),\nonumber\\
\mathcal{H}^\mathrm{c}_{i,j} &= \{\mathbf{x} \in \mathbb{R}^{dN}: h^\mathrm{c}_{i,j}(\mathbf{x}) \geq 0 \} \label{eq:h_conn}
\end{align}
% \end{footnotesize}

Then for the entire team with any given connectivity spanning graph $\mathcal{G}^\mathrm{c}=(\mathcal{V},\mathcal{E}^\mathrm{c})\subseteq\mathcal{G}$ to enforce, the desired set for safety and required connectivity are thus defined as:

% \begin{footnotesize}
\begin{align} \begin{aligned}\label{eq:H_safe_H_conn}
%\label{eq5}
    \mathcal{H}^\mathrm{s}&= \bigcap_{\{v_i,v_j \in \mathcal{V}:i>j\}} \!\mathcal{H}^\mathrm{s}_{i,j},\\
    \mathcal{H}^\mathrm{obs} &= \bigcap_{\{\forall i,o\}}\mathcal{H}^\mathrm{obs}_{i,o}, \;
    \mathcal{H}^\mathrm{c}(\mathcal{G}^\mathrm{c})= \bigcap_{(v_i,v_j)\in\mathcal{E}^\mathrm{c}} \!\mathcal{H}^\mathrm{c}_{i,j} 
\end{aligned}\end{align}
% \end{footnotesize}
% And the set of joint robot state $x$ satisfying both safety and connectivity for the robot team becomes\wl{we may not need H s plus c and B s plus c}
% \begin{align}\begin{aligned}\label{eq:H_safe_plus_conn}
% %\label{eq4}
%  \mathcal{H}^{s+c}(\mathcal{G}^{c}) = \mathcal{H}^{s} \cap \mathcal{H}^{c}(\mathcal{G}^{c})
% \end{aligned}\end{align}
% \wl{Yupeng, in the next time please do not use eq4, eq5, or fig1, fig2 as the labels for the equations. Use something like eq control barrier function as label name will make it much easier to remember and avoid conflicts when you put different equations into one paper later. For example, in your survey paper putting equations from different papers will have the same label eq5 for multiple equations even if they are different.}
Control Barrier Functions (CBF) \cite{ames2019control} are often used to define an admissible control space for robots rendering the desired set \textit{forward invariant}. The results are below.
\begin{lemma}
\label{Lemma 1}\label{lem:cbf}[summarized from \cite{ames2019control}]
Given a dynamical system affine in control and 
 % \yy{Given any Lipschitz continuous controller} and 
a desired set $\mathcal{H}$ as the 0-superlevel set of a continuous differentiable function $h(\mathbf{x}) : \mathcal{X} \rightarrow \mathbb{R}$, the function $h$ is called a control barrier function, if there exists an extended class-$\mathcal{K}$ function $\kappa(\cdot)$ such that $\sup_{\mathbf{u}\in\mathcal{U}}\{\dot{h}(\mathbf{x},\mathbf{u})+\kappa(h(\mathbf{x}))\}\geq 0$ for all $\mathbf{x}$. 
Any Lipschitz continuous controller $\mathbf{u}$ in the admissible control space $\mathcal{B}(\mathbf{x})$ rendering $\mathcal{H}$ forward invariant (i.e. keeping the system state $\mathbf{x}$ staying in $\mathcal{H}$ over time) thus becomes:%
% {\footnotesize
\begin{align}\label{eq:cbc_lemma}
    \mathcal{B}(\mathbf{x}) = \{ \mathbf{u}\in \mathcal{U} | \dot{h}(\mathbf{x},\mathbf{u}) + \kappa(h(\mathbf{x}))\geq 0 \}%, x \in \mathcal{X}
\end{align}
% }%
%\yy{where $\dot{h}(\mathbf{x},\mathbf{u}) = \frac{d}{dt}h(\mathbf{x})=\frac{\partial h(\mathbf{x})}{\partial\mathbf{x}}\dot{\mathbf{x}} = \frac{\partial h(\mathbf{x})}{\mathbf{x}}(f(\mathbf{x})+g(\mathbf{x})\mathbf{u})$ is the time derivative of $h(\mathbf{x})$.}
\end{lemma}
With this, the admissible control space for robots to stay collision-free and connected with maintained edges in a specified graph $\mathcal{G}^\mathrm{c}$ (i.e. $\mathbf{x}\in \mathcal{H}^\mathrm{c}(\mathcal{G}^\mathrm{c})$) in (\ref{eq:H_safe_H_conn}) are introduced by \cite{wang2016multi,wang2017safety} as follows (also referred as safety barrier certificates (SBC) and connectivity barrier certificates (CBC)):
% It is proved in \cite{ames2019control} that any controller $u\in\mathcal{B}(x)$ will render the safe state set $\mathcal{H}$ forward-invariant, i.e., if the system (\ref{eq:nonlinear}) starts inside the set $\mathcal{H}$ with $x(t=0)\in \mathcal{H}$, then it implies $x(t)\in\mathcal{H}$ for all $t>0$ under controller $u\in\mathcal{B}(x)$.
% \cite{wang2016multi,wang2017safety} proposed the safety barrier certificates (SBC) ${B}^s(\mathbf{x})$ and connectivity barrier certificates (CBC) ${B}^c(\mathbf{x},\mathcal{G}^c)$ using control barrier functions $h^s_{i,j}(\cdot)$ and $h^c_{i,j}(\cdot)$ that map the constrained safety and connectivity set Eq.~(\ref{eq2}), (\ref{eq3}) of $x$ to the admissible joint control space $u\in \mathbb{R}^{2N}$ for ensuring $h^s_{i,j}(\cdot), h^c_{i,j}(\cdot)\geq 0$ at all time. The result is summarized as follows.
% Therefore, the Safety Barrier Certificates (SBC) \cite{wang2017safety} and Connectivity Barrier Certificate(CBC) \cite{wang2016multi} are:

\vspace{-10pt}
% \begin{footnotesize}
\begin{align}
   & \mathcal{B}^\mathrm{s}(\mathbf{x}) = \{ \mathbf{u}\in \mathbb{R}^{qN} : \dot{h}^\mathrm{s}_{i,j}(\mathbf{x},\mathbf{u}) + \gamma h^\mathrm{s}_{i,j}(\mathbf{x})\geq 0, \forall i > j \} \label{eq7}\\
   & \mathcal{B}^\mathrm{obs}(\mathbf{x},\mathbf{x}^\mathrm{obs}) = \{ \mathbf{u}\in \mathbb{R}^{qN} : \nonumber \\
   & \qquad \qquad 
   \dot{h}^\mathrm{obs}_{i,o}(\mathbf{x},\mathbf{x}^\mathrm{obs},\mathbf{u}) + \gamma h^\mathrm{obs}_{i,o}(\mathbf{x},\mathbf{x}^\mathrm{obs})\geq 0, \forall i,o \}\label{bobs}\\
&\mathcal{B}^\mathrm{c}(\mathbf{x},\mathcal{G}^\mathrm{c})=\{\mathbf{u}\in \mathbb{R}^{qN}: \nonumber \\
& \qquad \qquad 
\dot{h}^\mathrm{c}_{i,j}(\mathbf{x},\mathbf{u})+\gamma h^\mathrm{c}_{i,j}(\mathbf{x})\geq 0, \forall (v_i,v_j)\in \mathcal{E}^\mathrm{c}\}\label{eq8}
\end{align}
% \end{footnotesize}

where $\gamma$ is a user-defined parameter in the particular choice of $\kappa(h(\mathbf{x}))=\gamma h(\mathbf{x})$ as in \cite{luo2020behavior}. It is proven in \cite{wang2016multi,wang2017safety} that the forward invariance of the safety set $\mathcal{H}^\mathrm{s}, \mathcal{H}^\mathrm{obs}$ and the connectivity set $\mathcal{H}^\mathrm{c}$ is ensured as long as the joint control input $\mathbf{u}$ stays in set $\mathcal{B}^\mathrm{s}(\mathbf{x})$, $\mathcal{B}^\mathrm{obs}(\mathbf{x},\mathbf{x}^\mathrm{obs})$, and $\mathcal{B}^\mathrm{c}(\mathbf{x},\mathcal{G}^\mathrm{c})$.
% In other words, the robots will always stay safe and connected if they are initially collision free and the control input lies in the set $\mathcal{B}^s(\mathbf{x})$. \wl{need to be revised} 
% The constrained control space in (\ref{eq:safebarrier}) corresponds to a class of linear constraints over pair-wise control inputs $u_i$ and $u_j$.
% Hence, the final admissible control space ensuring safety and connectivity (i.e. keeping $\mathcal{G}^c$ connected) in (\ref{eq:H_safe_plus_conn}) can be defined as the intersection of (\ref{eq7}) and (\ref{eq8}) as follows:\wl{we may not need H s plus c and B s plus c}
% \begin{align}\small\begin{aligned}
% \label{eq9}
%   \mathcal{B}^{s+c}(x, \mathcal{G}^c) = \mathcal{B}^{s}(x) \cap \mathcal{B}^{c}(x, \mathcal{G}^c)
% \end{aligned}\end{align}

\subsection{Line-of-Sight (LOS) Connectivity Constraints}\label{section32}
% \wl{we can give definition of los global and subgroup connectivity graph here first. }
% \noindent
Similarly, the desired set with pairwise LOS connectivity for robots $i,j$ can be described as follows.
% \begin{footnotesize}
\begin{align}\begin{split}
\label{hlos1}
    \mathcal{H}^\mathrm{los}_{i,j} = &\{\mathbf{x} \in \mathbb{R}^{dN}: h^\mathrm{los}_{i,j}(\mathbf{x},\mathcal{C}_\text{obs})\geq 0 \}=\\
    &\{\mathbf{x} \in \mathbb{R}^{dN}: \mathbf{x}_i(1-\beta)+\mathbf{x}_j\beta \notin \mathcal{C}_{\text{obs}},\;\forall \beta\in [0,1]\}
\end{split}\end{align}
% \end{footnotesize}

% Unlike the safety and connectivity constraints with straightforward forms of $h^\mathrm{s}_{i,j},h^\mathrm{c}_{i,j}$ in (\ref{eq:h_safe}) and (\ref{eq:h_conn}), 
Note that
it is non-trivial to define an analytical form of $h^\mathrm{los}_{i,j}$ w.r.t. $\mathbf{x}, \mathcal{C}_\mathrm{obs}$. In Section~\ref{Definition}, we will reformulate (\ref{hlos1}) and propose an explicit form of the continuously differentiable function $h^\mathrm{los}_{i,j}$ for $\mathcal{H}^\mathrm{los}_{i,j}$. Then for any LOS connectivity spanning graph $\mathcal{G}^\mathrm{slos}=(\mathcal{V},\mathcal{E}^\mathrm{slos})\subseteq\mathcal{G}^\mathrm{los}$ to enforce, the desired set $\mathbf{x}$ with the required LOS connectivity becomes:%

\vspace{-10pt}
% {\footnotesize
\begin{align}
\label{hlos}
    \mathcal{H}^\mathrm{los}(\mathcal{G}^\mathrm{slos})= \Big(\bigcap_{\{v_i,v_j \in \mathcal{V}:(v_i,v_j)\in\mathcal{E}^\mathrm{slos}\}} \mathcal{H}^\mathrm{los}_{i,j} \Big)\bigcap \mathcal{H}^\mathrm{c}(\mathcal{G}^\mathrm{slos})
\end{align}
% }%

Following Lemma~\ref{lem:cbf}, we develop the Line-of-Sight connectivity barrier certificates (LOS-CBC) $\mathcal{B}^\mathrm{los}(\mathbf{x},\mathcal{C}_\mathrm{obs},\mathcal{G}^\mathrm{slos})$ as follows (See Section~\ref{Definition} for detailed discussion) to characterize the admissible control space rendering $\mathcal{H}^\mathrm{los}(\mathcal{G}^\mathrm{slos})$ in (\ref{hlos}) forward invariant, i.e. satisfying both connectivity constraints with maintained edges $\mathcal{E}^\mathrm{slos}$ in $ \mathcal{G}^\mathrm{slos}$ and the occlusion-free condition:

\vspace{-15pt}
% {\footnotesize 
\begin{align}\begin{split}\label{eq:los_cbc_general}
% \footnotesize
&\mathcal{B}^\mathrm{los}(\mathbf{x},\mathcal{C}_\mathrm{obs},\mathcal{G}^\mathrm{slos})= \mathcal{B}^c(\mathbf{x},\mathcal{G}^\mathrm{slos})\bigcap \\
&\quad \{\mathbf{u}\in \mathbb{R}^{qN}: \\
   & \quad 
\dot{h}^\mathrm{los}_{i,j}(\mathbf{x}, \mathcal{C}_\mathrm{obs}, \mathbf{u})+\gamma h^\mathrm{los}_{i,j}(\mathbf{x}, \mathcal{C}_\mathrm{obs})\geq 0,\forall (v_i,v_j)\in \mathcal{E}^\mathrm{slos}\}
\end{split}\end{align}
\section{Problem Statement}
% To ensure global and subgroup are LOS connected, the global team and different sub-group team should stay LOS connected while performing the tasks.\wl{a non-sense sentence. We can just remove this sentence. I pointed this out so that we know how to avoid similar mistakes in the future} 
To simplify the discussion, for the rest of the paper we consider the single-integrator robot dynamics $\dot{\mathbf{x}}_{i} = \mathbf{u}_{i}$ with $\mathbf{u}_{i} \in \mathbb{R}^{d}$ (i.e. $q=d$) as commonly used in \cite{cavorsimulti,capelli2020connectivity}.
% (see appendix \cite{yupeng} that extends our results to more general control-affine systems). 
% Each robot state $\mathbf{x}_{i}$ and the discretized obstacle state $\mathbf{x}^\mathrm{obs}_{o}$ are the same as we described in the Section~\ref{section2}.
We assume the robotic team $\mathcal{S}$ with its real-time LOS communication graph $\mathcal{G}^\mathrm{los}$ has been assigned $M$ simultaneous tasks ($M\leq N$) with $M$ divided sub-groups $\mathcal{S}=\{\mathcal{S}_1,\ldots,\mathcal{S}_M\}$, where each robot $i$ has already been tasked to a sub-group $\mathcal{S}_m$ with a nominal task-related controller $\mathbf{u}_i=\mathbf{\hat{u}}_i\in \mathbb{R}^d$.

\noindent\textbf{Global LOS Connectivity}: A graph $\mathcal{G}^\mathrm{los}$ is said to be \textit{LOS connected} if there is at least one occlusion-free path between every pair of vertices on the graph.

\noindent\textbf{Subgroup LOS Connectivity}: A graph $\mathcal{G}^\mathrm{los}$ is said to be \textit{Subgroup LOS connected} if there is at least one occlusion-free path between every pair of vertices in each induced los subgroup graph $\mathcal{G}^\mathrm{los}_m = \mathcal{G}^\mathrm{los}[\mathcal{V}_m]\subseteq\mathcal{G}^\mathrm{los}, \forall m=1,\ldots,M$, where $\mathcal{V}_m\subseteq \mathcal{V}$ contains all robots within the same sub-group $\mathcal{S}_m$.

To ensure smooth information exchange both globally and locally, the LOS communication graph $\mathcal{G}^\mathrm{los}$ containing all the robots should satisfy both global and subgroup LOS connectivity at all time.
We assume the global and subgroup connectivity of the LOS communication graph $\mathcal{G}^\mathrm{los}$ are satisfied initially. 
Hence, the step-wise optimization problem
% of minimally constrained multi-robot coordination with global and subgroup LOS connectivity maintenance 
is defined as follows:
\vspace{-0.5cm}

% {\footnotesize
\begin{align}
 &\mathbf{u}^* = \argmin_{\mathcal{G}^\mathrm{slos},\mathbf{u}} \sum_{i=1}^{N}||\mathbf{u}_i-\mathbf{\hat{u}}_i||^2 \label{eq:rawobj}\\
 \text{s.t.} &\quad \mathcal{G}^\mathrm{slos}=(\mathcal{V},\mathcal{E}^\mathrm{slos})\subseteq \mathcal{G}^\mathrm{los}\quad \text{is LOS connected} \label{eq:rawglobal}\\
&\quad  \mathcal{G}^\mathrm{slos}_m=\mathcal{G}^\mathrm{slos}[\mathcal{V}_m]\quad \text{is LOS connected},\forall m=1,\ldots,M \label{eq:rawconn}\\
&\quad \mathbf{u}\in \mathcal{B}^\mathrm{s}(\mathbf{x})\bigcap\mathcal{B}^\mathrm{obs}(\mathbf{x},\mathbf{x}^\mathrm{obs})\bigcap \mathcal{B}^\mathrm{los}(\mathbf{x},\mathcal{C}_\mathrm{obs},\mathcal{G}^\mathrm{slos})\label{eq:rawconst}\\
&\quad ||\mathbf{u}_i|| \leq u_\mathrm{max},\forall i=1,\ldots,N \notag
\end{align}
% }%

%\wl{Here we may need to iterate more to specify the relation between the los global and subgroup connectivity graphs, the satisfying subgraph, and the optimal satisfying subgraph.} 
The optimization problem (\ref{eq:rawobj}) seeks to minimally modify the given nominal task-related controller $\mathbf{\hat{u}}_i$ for each robot $i$ at each time step, while respecting the required LOS connectivity constraints
determined by a satisfying $\mathcal{G}^\mathrm{slos}\subseteq \mathcal{G}^\mathrm{los}$. 
Note that the LOS communication graph $\mathcal{G}^\mathrm{los}$ at each time step could have multiple subgraphs $\{\mathcal{G}^\mathrm{slos}\}$ that satisfy global and subgroup LOS connectivity, each of which defines a specific set of control constraints by $\mathcal{B}^\mathrm{los}(\mathbf{x},\mathcal{C}_\mathrm{obs},\mathcal{G}^\mathrm{slos})$ in (\ref{eq:rawconst}). Thus 
% instead of keeping all edges LOS connected in $\mathcal{G}$ that incur more LOS constraints, 
it is beneficial to select an optimal subgraph $\mathcal{G}^\mathrm{slos*}\subseteq \mathcal{G}^\mathrm{los}$ 
% (that is both global and subgroup LOS connected) 
to generate the corresponding LOS-CBC $\mathcal{B}^\mathrm{los}(\mathbf{x},\mathcal{C}_\mathrm{obs},\mathcal{G}^\mathrm{slos*})$ in (\ref{eq:rawconst}), so that (a) it specifies the least number of edges $|\mathcal{E}^\mathrm{slos*}|$ to maintain (i.e. less LOS constraints needed to ensure global and subgroup LOS connectivity), and (b) enforcing constraints $\mathcal{B}^\mathrm{los}(\mathbf{x},\mathcal{C}_\mathrm{obs},\mathcal{G}^\mathrm{slos*})$ would introduce minimum control deviation from $\mathbf{\hat{u}}_i$ for all $i$ (i.e. least constraining $\mathbf{\hat{u}}_i$ compared to maintain other $\mathcal{G}^\mathrm{slos}$). 
Therefore, the problem can be considered as a bilevel optimization process: (i) find the optimal spanning subgraph $\mathcal{G}^\mathrm{slos*} \subseteq \mathcal{G}^\mathrm{los}$ to preserve, i.e. one satisfying (\ref{eq:rawglobal}),(\ref{eq:rawconn}) and introducing least constraining LOS constraints for keeping all edges in $\mathcal{E}^\mathrm{slos*}$ LOS connected, and (ii) %\yy{solve the Quadratic Programming(QP) to }obtain step-wise control $\mathbf{u}^{*} \in \mathbb{R}^{q\times N}$ restricted by the maximum value $\mathbf{u}_{max}$ and minimally deviated from $\mathbf{\hat{u}}_i$ subject to \yy{linear} constraints (\ref{eq:rawconst})  with $\mathcal{G}^\mathrm{los} =\mathcal{G}^\mathrm{los*}$. 
use $\mathcal{G}^\mathrm{slos*}$ to define control constraints $\mathcal{B}^\mathrm{los}(\mathbf{x},\mathcal{C}_\mathrm{obs},\mathcal{G}^\mathrm{slos*})$ and solve the Quadratic Programming(QP) problem to obtain step-wise control $\mathbf{u}^{*} \in \mathbb{R}^{dN}$.
% , which is minimally deviated from $\mathbf{\hat{u}}_i$.
% subject to linear control constraints in  (\ref{eq:rawconst}).
As the optimization process ($\ref{eq:rawobj}$) is repeatedly solved at every time step, the selected $\mathcal{G}^\mathrm{slos*} \subseteq \mathcal{G}^\mathrm{los}$ will be re-computed overtime 
% % \wl{and the solution of G always exists according to our theorem xx} 
as $\mathcal{G}^\mathrm{los}$ changes due to updated robot positions, thus enabling flexible coordination with dynamic LOS connectivity maintenance.

\section{Method}
\subsection{Line-of-Sight Connectivity Barrier Certificates}\label{Definition}
%The brief introduction of Line-of-Sight Connectivity Barrier Certificate (LOS-CBC) is given in Section~\ref{section32}, while how to compose such $h_{i,j,o}^\mathrm{los}$ for $\mathcal{B}^\mathrm{los}(\mathbf{x},\mathcal{C}_\mathrm{obs},\mathcal{G}^\mathrm{los})$ to define the admissible control space is challenging. 
%In this section, we will introduce the composition of $h_{i,j,o}^\mathrm{los}$ to characterize the occlusion-free LOS condition given pairwise robots positions $\mathbf{x}_i,\mathbf{x}_j$ and the state of discreted points $\mathbf{x}^\mathrm{obs}_{o}$.

Since it is non-trivial to derive an analytical form of $h^\mathrm{los}_{i,j}(\mathbf{x},\mathcal{C}_\mathrm{obs})$ based on the occlusion-free condition in (\ref{hlos1}), 
% check whether the infinite segment points on the pair-wise robot communication edge are free from occlusions by all the obstacles or not,
%The high level idea is that for any pairwise robots $i,j$ in line-of-sight,
here we propose to use an approximation method such as an ellipsoid to 
represent a LOS communication edge $(v_i,v_j)\in\mathcal{E}^\mathrm{los}$ and to analytically determine 
% the current LOS communication edge $(v_i,v_j)\in\mathcal{E}^\mathrm{los}$ to maintain, so that the ellipsoid representation can be leveraged to determine
whether any obstacle is intersecting with this edge. To prevent
% the ellipsoidal approximation to lead to 
overly conservative approximation, we formulate the ellipsoidal approximation as a Minimum Volume Enclosing Ellipsoid (MVEE) problem \cite{moshtagh2005minimum} covering the edge $(v_i,v_j)\in\mathcal{E}^\mathrm{los}$.

As proved in \cite{todd2007khachiyan}, a set of $2d$ points in space $\mathbb{R}^d$ suffice to determine a MVEE that tightly covers those points. To ensure a good approximation, for each edge $(v_i,v_j)\in\mathcal{E}^\mathrm{los}$ we use a particular choice of $2d$ points analytically defined as $\mathcal{P}_{i,j}= \{\mathbf{p}^{1}_{i,j},...,\mathbf{p}^{2d}_{i,j}\}$ to reconstruct the corresponding MVEE covering the line segment $\mathbf{x}_i(1-\beta)+\mathbf{x}_j\beta ,\;\forall \beta\in [0,1]$. $\mathbf{p}^{1}_{i,j}=\mathbf{x}_i,
\mathbf{p}^{2}_{i,j}=\mathbf{x}_j$ are two vertices on the major principle axis of the MVEE and $\{\mathbf{p}^{3}_{i,j},...,\mathbf{p}^{2d}_{i,j}\}$ are the other vertices on the non-major principle axis of the MVEE centered at the middle point $\mathbf{p}^{0}_{i,j} = \frac{\mathbf{x}_i+\mathbf{x}_j}{2}$ with the same semi-axis length as $||\mathbf{p}^{l}_{i,j}-\mathbf{p}^{0}_{i,j}||=\delta,\forall l=3,\ldots,2d $, a small value determined beforehand to reflect the "thickness" of the ellipsoid. With that, 
% this is analytically computed so efficient.
% the form of the ellipsoid
% closely around the pairwise communication edge between robots $i,j$ for fitting the MVEE, where $z = \frac{d\times(d+3)}{2}$.
%To be specific, the sampled fix-size point set contains the state of pair-wise robot $i$ and $j$, the state of the communication edge's midpoint(i.e $\frac{\mathbf{x}_i +\mathbf{x}_j}{2}$), and $\frac{d*(d+3)}2-3$ number of points sampled around the communication edge's midpoint. 
the corresponding MVEE \cite{todd2007khachiyan} for edge $(v_i,v_j)\in\mathcal{E}^\mathrm{los}$ is determined by a positive-definite matrix $Q_{i,j}\in \mathbb{R}^{d\times d}$ and centered at $\mathbf{p}^{0}_{i,j}$, where 
$Q_{i,j}\leftarrow \argmin \text{det} (Q_{i,j}^{-1})$ and $(\mathbf{p}^{r}_{i,j}-\mathbf{p}^{0}_{i,j})^{T}Q_{i,j}(\mathbf{p}^{r}_{i,j}-\mathbf{p}^{0}_{i,j})\leq1,r=1,...,2d$.
% describing the shape of ellipsoidal approximation around a set of points centered at the communication edge's midpoint.
% The detailed discussion related to the size of the sampled point set and how to sample the points around the communication edge's midpoint will be presented in the Appendix.
Note that the approximated ellipsoid characterized by $Q_{i,j}$ will update as robots $i,j$ move over time. 
% As our goal is to describe the intersection condition of obstacles and the communication edge between the pairwise robots, for the polytopic obstacles in the environment, the boundaries of each obstacles are modeled as a series of rigid spheres to ensure real-time computation efficiency. It is a legit way to model the presence of obstacles, as if $\mathcal{C}_\text{obs}$ overlaps with the a communication edge, there must be points on at least one boundary is within the ellipsoid.
% In this work  The positions of rigid spheres representing obstacle $\mathcal{O}=\{1,\ldots,K\}$ is denoted as $x_o\in \mathbb{R}^2, o\in \mathcal{O}$.
Then the function of $h_{i,j}^\mathrm{los}$ for occlusion-free condition and its 0-superlevel set $\mathcal{H}_{i,j}^\mathrm{los}$ in (\ref{hlos1}) are analytically re-defined as, $\forall o$:

% \vspace{-12pt}
% \begin{footnotesize}
\begin{equation}\label{los1}
    \begin{split}
        &h_{i,j,o}^\mathrm{los}(\mathbf{x},\mathbf{x}^\mathrm{obs}) = (\mathbf{x}^\mathrm{obs}_o-\mathbf{p}^{0}_{i,j})^{T}Q_{i,j}(\mathbf{x}^{\mathrm{obs}}_o-\mathbf{p}^{0}_{i,j})-1, \\
        & \qquad \qquad \qquad \qquad \qquad \qquad \qquad \qquad \qquad 
        \forall (v_i,v_j)\in\mathcal{E}^\mathrm{los}\\
       &\mathcal{H}_{i,j,o}^\mathrm{los}=\{{\mathbf{x}}\in \mathbb{R}^{d N}:h^\mathrm{los}_{i,j,o}(\mathbf{x},\mathbf{x}^\mathrm{obs})\geq 0\;,\; \forall (v_i,v_j)\in \mathcal{E}^\mathrm{los} ,\forall o\} \\
       &\mathcal{H}_{i,j}^\mathrm{los} = \bigcap_{\forall o} \mathcal{H}^\mathrm{los}_{i,j,o}
    \end{split}
    \vspace{-10pt}
\end{equation}
% \end{footnotesize}
% \vspace{-10pt}

%where $o\in \mathbb{R}^d$ represents any point on the boundary of the obstacle area $\mathcal{C}_\mathrm{obs}$. 
%Equation~(\ref{los1}) ensures robots remain occlusion-free from the obstacles as long as they stay in set $\mathcal{H}_{i,j,o}^\mathrm{los}$.
% the LOS-CBC will always keep the communication edges in MAS from occlusion disruption.
%\wl{los cbc should be formally defined here, following the lemma of control barrier function in the preliminary section, so that an explicit condition specifying admissible control space for ensuring los connectivity between pairwise robots i and j is given.}
% Given this reformulated Line-of-Sight connectivity constraint, 
Following Lemma~\ref{lem:cbf} with control barrier functions, we now formally define the Line-of-Sight Connectivity Barrier Certificates (LOS-CBC) as follows.
% \wl{correct other places so that we have LOS \textbf{connectivity} barrier certificates instead of LOS \textbf{control} barrier certificate} 
% describing admissible control space for the entire robotic team to render $\mathcal{H}^{los}(\mathcal{G}^{los})$ in (\ref{hlos}) forward invariant.
% \wl{check if we have described what is forward invariant earlier.}\yy{yes we have}
\begin{lemma}
\label{loscbcdefinition}
\textbf{Line-of-Sight Connectivity Barrier Certificates(LOS-CBC):} 
Given a LOS communication spanning graph $\mathcal{G}^\mathrm{slos}=(\mathcal{V},\mathcal{E}^\mathrm{slos})\subseteq \mathcal{G}^\mathrm{los}$ and a desired set $\mathcal{H}^\mathrm{los}(\mathcal{G}^\mathrm{slos})$ in (\ref{hlos}) with $h^\mathrm{los}_{i,j,o}$ from (\ref{los1}), for any Lipschitz continuous controller 
$\mathbf{u}$, the Line-of-Sight connectivity barrier certificates (LOS-CBC) as the admissible control space $ \mathcal{B}^\mathrm{los}(\mathbf{x},\mathcal{C}_\mathrm{obs},\mathcal{G}^\mathrm{slos})$ defined below renders $\mathcal{H}^\mathrm{los}(\mathcal{G}^\mathrm{slos})$ forward invariant (i.e keeping joint robot state staying in $\mathcal{H}^\mathrm{los}(\mathcal{G}^\mathrm{slos})$):%
\vspace{-5pt}
% {\footnotesize
\begin{align}
\begin{split}\label{eq:los_cbc_definition}
&\mathcal{B}^\mathrm{los}(\mathbf{x},\mathcal{C}_\mathrm{obs},\mathcal{G}^\mathrm{slos})= \mathcal{B}^c(\mathbf{x},\mathcal{G}^\mathrm{slos})\bigcap \{\mathbf{u}\in \mathbb{R}^{dN}:\\
&\;\dot{h}^\mathrm{los}_{i,j,o}(\mathbf{x}, \mathbf{x}^\mathrm{obs}, \mathbf{u})+\gamma h^\mathrm{los}_{i,j,o}(\mathbf{x},\mathbf{x}^\mathrm{obs})\geq 0,\forall (v_i,v_j)\in \mathcal{E}^\mathrm{slos},\forall o\}%\in \partial{\mathcal{C}_\mathrm{obs}}\}
\end{split}
\end{align}
% }%
where $\dot{h}_{i,j,o}^\mathrm{los}(\mathbf{x},\mathbf{x}^\mathrm{obs}, \mathbf{u}) = -(\mathbf{x}^\mathrm{obs}_{o}-\frac{\mathbf{x}_i+\mathbf{x}_j}{2})^{T}Q_{i,j}(\mathbf{u}_{i}+\mathbf{u}_{j}).$ 
\end{lemma}%
$\mathcal{B}^\mathrm{los}(\mathbf{x},\mathcal{C}_\mathrm{obs},\mathcal{G}^\mathrm{slos})$ relies on the composition of $h_{i,j}^c$ and $h_{i,j,o}^\mathrm{los}$. 
% \yy{
Proof of Lemma~\ref{loscbcdefinition} is provided in section~\ref{app:sec:los-cbc}.
\subsection{Minimum Line of Sight Connectivity Constraint Spanning Tree (MLCCST)}
% Given the bilevel optimization problem (\ref{eq:rawobj}), now consider the first sub-problem of finding the optimal LOS communication graph $\mathcal{G}^\mathrm{los*}\subseteq \mathcal{G}$.
%^so that by enforcing $\mathbf{u}\in \mathcal{B}^\mathrm{los}(\mathbf{x},\mathcal{C}_\mathrm{obs},\mathcal{G}^\mathrm{los*})$, we have that (a) the robotic team always satisfy global and subgroup LOS connectivity in (\ref{eq:rawglobal}) and (\ref{eq:rawconn}), and (b) $\mathbf{u}\in \mathcal{B}^\mathrm{los}(\mathbf{x},\mathcal{C}_\mathrm{obs},\mathcal{G}^\mathrm{los*})$ introduces minimum constraints over the nominal joint multi-robot controller. 
% used for determining minimally safety and Line-of-Sight connectivity constraints in the form of (\ref{eq7}) and (\ref{eq:los_cbc_definition}) over the nominal multi-robot controllers. 
As each edge $(v_i,v_j)\in \mathcal{E}^\mathrm{slos}$ in a candidate graph $\mathcal{G}^\mathrm{slos}$ enforces one particular LOS connectivity requirement restricting the motion of robot $i$ and $j$, the desired graph $\mathcal{G}^\mathrm{slos*}$ whose edges define the minimum LOS connectivity constraints (i.e. most \textit{unlikely} to be violated if following nominal control $\mathbf{\hat{u}}_i,\mathbf{\hat{u}}_j$) must exist among the set of all spanning trees $\mathcal{T}^\mathrm{los}$ of $\mathcal{G}^\mathrm{los}$ that have the minimum number of edges (i.e. $N-1$) for $\mathcal{G}^\mathrm{slos*}$ to stay LOS connected.
Hence, the first sub-problem boils down to finding the optimal spanning tree $\mathcal{G}^\mathrm{slos*}=\mathcal{T}^\mathrm{los*}\subseteq \mathcal{G}^\mathrm{los}$ whose edges invoke the minimum LOS connectivity constraints in the form of (\ref{eq:los_cbc_definition}) over the nominal robots' controller.
Below we introduce the connectivity weight $w^\mathrm{d}_{i,j}$, line of sight weight $w^\mathrm{los}_{i,j}$, and the edge weight $w^\mathrm{d+los}_{i,j}$ as the sum of the two to heuristically quantify how \textit{unlikely} the pairwise LOS constraints are to be violated under the nominal controller $\mathrm{\hat{u}}$, $\forall (v_i,v_j)\in\mathcal{E}^\mathrm{los}$ :
% \vspace{-0.2cm}
% {\footnotesize
\begin{align}
w^\mathrm{d}_{i,j} &=  \dot{h}^\mathrm{c}_{i,j}(\mathbf{x},\mathbf{\Hat{u}})+\gamma h^\mathrm{c}_{i,j}(\mathbf{x})\label{wdweight}\\
w^\mathrm{los}_{i,j} &=\frac{1}{F}\sum_{o=1}^{F}({{\dot{h}_{i,j,o}^\mathrm{los}(\mathbf{x},\mathbf{x}^\mathrm{obs},\mathbf{\hat{u}})+\gamma h_{i,j,o}^\mathrm{los}(\mathbf{x},\mathbf{x}^\mathrm{obs})}}) \label{locweight} \\
w^\mathrm{d+los}_{i,j} &= \left\{ \begin{gathered} w^\mathrm{los}_{i,j} + w^\mathrm{d}_{i,j} \quad \text{if}\quad \forall o , \quad h_{i,j,o}^\mathrm{los}(\mathbf{x},\mathbf{x}^\mathrm{obs}) \geq 0\\%\quad 
   \epsilon  \quad \text{otherwise, namely} \quad \exists o, \quad h_{i,j,o}^\mathrm{los}(\mathbf{x},\mathbf{x}^\mathrm{obs}) < 0,\label{WLOSdefine}
   \end{gathered}\right.
\end{align}
% }%

where $\epsilon \in \{\epsilon \ll 0: \epsilon \ll  w^\mathrm{d+los}_{i,j},\forall (v_i,v_j)\in\mathcal{E}\}$ is a unique user-defined constant for the entire graph $\mathcal{G}^\mathrm{los}$. $w^\mathrm{d}_{i,j}, w^\mathrm{los}_{i,j}$ indicate the level of violation of the two constraints between robots $i,j$ under the nominal controller $\mathbf{\hat{u}}_i, \mathbf{\hat{u}}_j$ from $\mathbf{\hat{u}}$, with the higher value of $w^\mathrm{d}_{i,j}, w^\mathrm{los}_{i,j}$ the less violated the constraints are. 
% \yy{sufficient condition delete next sentence}
In particular, $w^\mathrm{los}_{i,j}$ defined in (\ref{locweight}) reflects preference over LOS edges that are unlikely to be violated on average w.r.t. all of the obstacles, thus implying a larger slackness in general when being preserved compared to others.
By introducing the penalization term of $\epsilon$ in (\ref{WLOSdefine}),
the Line-of-Sight connectivity weight $w^\mathrm{d+los}_{i,j}$ naturally reveals those LOS connectivity edges $(v_i,v_j) \in \mathcal{E}^\mathrm{los}$ satisfying our redefined occlusion-free condition (\ref{los1}). 
% also naturally distinguishes whether the edges $(v_i,v_j) \in \mathcal{E}$ in graph $\mathcal{G}$ is a LOS connectivity edge $(v_i,v_j) \in \mathcal{E}^\mathrm{los}$ free from occlusions. 
% With that, Lemma~\ref{guarantee} will prove that the optimal spanning tree is the sub-graph of the LOS connectivity graph (i.e. $\mathcal{G}^{los*}\in\mathcal{G}^{los}$). In summary, the Line-of-Sight weight ensures each candidate spanning tree $\mathcal{T}^{los}\in\mathcal{G}^{los}$. \yy{rewrite}
Note that to further reduce computation burden in (\ref{locweight}) with a large number of $F$ discretized obstacles, tools such as efficient nearest neighbor searches \cite{atramentov2002efficient} could be employed to only consider a smaller number of obstacles closer to the robots $i,j$, which are beyond the scope of this paper.
With that, each candidate spanning tree $\mathcal{T}^\mathrm{los}\subseteq \mathcal{G}^\mathrm{los}$ 
% $\mathcal{T}^{los}\in\mathcal{G}^{los}\in\mathcal{G}$ 
is redefined as a weighted spanning tree
$\mathcal{T}^\mathrm{los}_w=(\mathcal{V},\mathcal{E}^{T},\mathcal{W}^T)$
% $\mathcal{T}_w^{los}=(\mathcal{V},\mathcal{E}^{T},\mathcal{W}^T)$ 
where $\mathcal{E}^T \subseteq \mathcal{E}^\mathrm{los}$ 
% $\mathcal{E}^T \in \mathcal{E}^{los}$
with weight $\mathcal{W}^T=\{-w^\mathrm{d+los}_{i,j}\}$. Hence the optimal LOS connectivity graph $\mathcal{G}^\mathrm{slos*}$ with constraints in (\ref{eq:rawglobal}),(\ref{eq:rawconn}) can be defined as follows. 
% \vspace{-0.2cm}
\begin{equation}%\footnotesize
\begin{split}
   \mathcal{G}^\mathrm{slos*}&=\!\argmax_{\{\mathcal{T}_w^\mathrm{los}\}}\sum_{(v_i,v_j)\in \mathcal{E}^{T}}w^\mathrm{d+los}_{i,j}\\
   &=\!\argmin_{\{\mathcal{T}_w^\mathrm{los}\}}\!\sum_{(v_i,v_j)\in \mathcal{E}^{T}}-w^\mathrm{d+los}_{i,j}\\
 \text{s.t.} &\quad  \mathcal{T}^\mathrm{los}_m=\mathcal{T}_w^\mathrm{los}[\mathcal{V}_m]\quad \text{is LOS connected},\forall m=1,\ldots,M \label{eq:expobj}  
\end{split}
\end{equation}

The optimal solution of (\ref{eq:expobj}) is the Minimal Spanning Tree (MST) of the LOS communication graph $\mathcal{G}^\mathrm{los}$ weighted by $\{-w^\mathrm{d+los}_{i,j}\}$ with additional subgroup LOS connectivity constraints. 
% In Lemma~\ref{guarantee} (Section~\ref{sec:theory_analysis}), the optimal spanning graph $\mathcal{G}^\mathrm{slos*}$ from (\ref{eq:expobj}) is proved to be a LOS connected graph. 
% To further address the subgroup LOS connectivity constraint, 
Now we define another class of spanning trees as follows and relate its unconstrained MST to the solution of the subgroup LOS connectivity constrained MST in (\ref{eq:expobj}).
\begin{definition}\label{def:lccst}
\textbf{Line of Sight Connectivity Constraint Spanning Tree(LCCST):}
Given a LOS communication graph $\mathcal{G}^\mathrm{los}=(\mathcal{V},\mathcal{E}^\mathrm{los})$
% weighted spanning tree $\mathcal{T}_w^c=(\mathcal{V},\mathcal{E}^T,\mathcal{W}^T)$ 
and for all edges $(v_i,v_j)\in \mathcal{E}^\mathrm{los}$ on $\mathcal{G}^\mathrm{los}$, redefine their weights by the following.
% {\footnotesize
\begin{align}
\label{eq:neww}
w'_{i,j} = \left\{ \begin{gathered}
  \lambda\cdot w^\mathrm{d+los}_{i,j}, \quad \text{if} \quad \text{$v_i$ and $v_j$ are in the same sub-group } \hfill \\
  w^\mathrm{d+los}_{i,j}, \quad \text{if} \quad \text{$v_i$ and $v_j$ are in different sub-groups} \hfill \\
\end{gathered}  \right.
\end{align}
% }%

where $\lambda\in\{\lambda \gg 1: \lambda \cdot w^\mathrm{d+los}_{i,j}\gg w^\mathrm{d+los}_{i',j'},\forall v_i,v_i',v_j,v_j' \in \mathcal{V}\}$ is a unique user-defined constant for the entire graph $\mathcal
{G}^\mathrm{los}$. The weight-modified graph from $\mathcal{G}^\mathrm{los}$ is thus denoted as $\mathcal{G}^\mathrm{los'}=(\mathcal{V},\mathcal{E}^{los},\mathcal{W}')$ with $\mathcal{W}'=\{-w'_{i,j}\}$. Then we call the redefined LOS spanning tree $\mathcal{T}^\mathrm{los'}_w=(\mathcal{V},\mathcal{E}^{T},\mathcal{W}^{T'})\subseteq \mathcal{G}^\mathrm{los'}$ as the Line of Sight Connectivity Constraint Spanning Tree(LCCST).
\end{definition}

% Finally, the calculation for the weight of the graph $\mathcal{G}^{'}$ is shown in Algorithm \ref{alg:weight}.
%The Definition \ref{def:lccst} introduces a new class of spanning trees (LCCST) $\{\mathcal{T}_w^\mathrm{los'}\}$ equivalent to the original LOS spanning trees $\{\mathcal{T}_w^\mathrm{los}\}$ with inflated weights over the edges connecting robots in the same subgroup. In another word, 
Considering all those LOS edges satisfying our redefined occlusion-free condition (\ref{los1}), the designed parameter $\lambda$ in (\ref{eq:neww}) ensures that the new weight $\{-w'_{i,j}\}$ over those edges connecting different subgroups are always larger than any edges within the same subgroup.
% (See Lemma~\ref{lemma:subgroup})
In this way, we present Theorem~\ref{theorem:mlccst} to transform the constrained MST problem in (\ref{eq:expobj}) into an unconstrained MST problem with the same optimally guarantee, ensuring the optimal MST computed from $\{\mathcal{T}^\mathrm{los'}_w\}$ contains the MST of each subgroup as well.

% \begin{algorithm}[t]\footnotesize
% \caption{Computation for the weight of graph $\mathcal{G}^{'}$}
%     \label{alg:weight}
%     \begin{algorithmic}[1]
%     \Input{$x_{i}$:the state of the robots, $u_i$:the dynamics of the robots, $\mathcal{C}_\text{obs}$ the space of the obstacles}
%     \Output{weight for each potential edge exist in the entire graph $\mathcal{G}'$}
%     \Function{EdgeWeight}{$x_i$,$x_j$,$u_i$,$u_j$,$\mathcal{C}_\text{obs}$}
%     \State $w^{d}_{i,j} =  \dot{h}^c_{i,j}(\mathbf{x_i},\mathbf{x_i},u_i,u_j)+h^c_{i,j}(\mathbf{x})$
%     \State $w^{los}_{i,j} =\dot{h}_{i,j}^{los}(x,\mathcal{C}_\text{obs})+\gamma h_{i,j}^{los}(x,\mathcal{C}_\text{obs})$
%     \If{${h}_{i,j}^{los}(x,\mathcal{C}_\text{obs})\geq 0$}
%     \State $w^{d+los}_{i,j} = w^{d}_{i,j}+w^{los}_{i,j}$
%     \Else
%     \State$w^{d+los}_{i,j} =\epsilon $
%     \EndIf
%     \If{\text{$v_i$ and $v_j$ are in the same sub-group }}
%     \State {$w'_{i,j}=\lambda\cdot w^{d+los}_{i,j}$}
%     \Else
%     \State {$w'_{i,j}=w^{d+los}_{i,j}$}
%     \EndIf
%     \State \Return{$-w'_{i,j}$}
%     \EndFunction
%     \end{algorithmic}
% \end{algorithm}
% \setlength{\textfloatsep}{0pt}
% \wl{why however here?? the following sentence tells readers nothing} To relax the constrained MST problem in (\ref{eq:neww}) into unconstrained MST problem, the Theorem \ref{theorem:mlccst} has been introduced

\begin{theorem}\label{theorem:mlccst}
Given the redefined Line of Sight Connectivity Constraint Spanning Tree (LCCST) $\mathcal{T}_w^\mathrm{los'}=(\mathcal{V},\mathcal{E}^{T},\mathcal{W}^{T'})$ in Definition \ref{def:lccst} and denote minimum weighted LCCST as $\bar{\mathcal{T}}_w^\mathrm{los'}=\argmin_{\{\mathcal{T}_w^\mathrm{los'}\}} \sum_{(v_i,v_j)\in \mathcal{E}^{T}}\{-w'_{i,j}\}$, then we have: $\mathcal{G}^\mathrm{slos*}=\bar{\mathcal{T}}_w^\mathrm{los'}$ in (\ref{eq:expobj}). Namely, the Minimum Spanning Tree $\bar{\mathcal{T}}_w^\mathrm{los'}$ of $\mathcal{G}^\mathrm{los'}$ is the optimal solution of $\mathcal{G}^\mathrm{slos*}$ in (\ref{eq:expobj}) and we call the graph $\bar{\mathcal{T}}_w^\mathrm{los'}$ as Minimum Line-of-Sight Connectivity Constraint Spanning Tree (MLCCST) of the original LOS communication graph $\mathcal{G}^\mathrm{los}$.
\end{theorem}
\vspace{-5pt}
The detailed proof is presented in Section~\ref{app:sec:mlccst_theorem}. Note that the MLCCST $\bar{\mathcal{T}}_w^\mathrm{los'}$ could be updated over time due to dynamically changing $\mathcal{G}^\mathrm{los}$ between each time step. 

Finally, our proposed MLCCST algorithm for computing the MLCCST $\bar{\mathcal{T}}_w^\mathrm{los'}$ and the revised multi-robot controllers is summarized in Algorithm~\ref{alg:Dynamic}. 
As shown by Theorem~\ref{theorem:mlccst}, the derived $\mathcal{G}^\mathrm{slos*}=\bar{\mathcal{T}}_w^\mathrm{los'}\subseteq \mathcal{G}^\mathrm{los}$ satisfies the LOS conditions in (\ref{eq:rawglobal}) and (\ref{eq:rawconn}). Recalling the weight definition in (\ref{WLOSdefine}), the \textit{minimally weighted} nature of the global and subgroup LOS connected $\bar{\mathcal{T}}_w^\mathrm{los'}$  thus indicates the resultant \emph{least constraining} LOS constraints $\mathcal{B}^\mathrm{los}(\mathrm{x},\mathcal{C}_\mathrm{obs},\bar{\mathcal{T}}_w^\mathrm{los'})$ given the task-related nominal control, and so to provide the greatest flexibility for the team. Hence, the computed $\mathbf{u}^*$ by Algorithm~\ref{alg:Dynamic} is a desired 
% optimal 
solution for problem~(\ref{eq:rawobj}).
Note that the control constraints $\mathbf{u}\in \mathcal{B}^\mathrm{s}(\mathbf{x})\bigcap \mathcal{B}^\mathrm{los}(\mathrm{x},\mathcal{C}_\mathrm{obs},\bar{\mathcal{T}}_w^\mathrm{los'})\bigcap\mathcal{B}^\mathrm{obs}(\mathbf{x},\mathbf{x}^\mathrm{obs})$ on Line~\ref{alg:line:find_u_star} (Algorithm~\ref{alg:Dynamic}) are linear w.r.t. $\mathbf{u}$, therefore making it a standard step-wise Quadratic Programming (QP) that could be efficiently solved in real time.
%Recall that the control constraints $\mathbf{u}\in \mathcal{B}^\mathrm{s}(\mathbf{x})\bigcap \mathcal{B}^\mathrm{los}(\mathrm{x},\mathcal{C}_\mathrm{obs},\bar{\mathcal{T}}_w^\mathrm{los'})$ on Line~\ref{alg:line:find_u_star} (Algorithm~\ref{alg:Dynamic}) are linear w.r.t. $\mathbf{u}$, hence making it a standard step-wise Quadratic Programming (QP) that could be efficiently solved in real time.

{\vspace{-5pt}
\begin{algorithm}
% \footnotesize{
\caption{MLCCST Algorithm}
    \label{alg:Dynamic}
    \begin{algorithmic}[1]
    \Input{$\mathbf{x}$-the current states (positions) of the robots, $\mathbf{\hat{u}}$-the nominal task-related multi-robot controller, $\mathcal{C}_\mathrm{obs}$ the occupied space of the obstacles}
    \Output{The desired minimally modified controller $\mathbf{u}^*\in\mathbb{R}^{dN}$ from (\ref{eq:rawobj}) }
    \Function{MLCCST}{$\mathbf{x}$, $\mathbf{\hat{u}}$, $\mathcal{C}_\mathrm{obs}$}
    \For{Each Time Step}
    \For { All Edges $(v_i,v_j)\in \mathcal{E}^\mathrm{los}$ of current LOS communication Graph $\mathcal{G}^\mathrm{los}=(\mathcal{V},\mathcal{E}^\mathrm{los})$}
     \State Weight assignment: $\mathcal{W}'_{i,j}$ $\gets$ $-w'_{i,j}$ using (\ref{WLOSdefine}) and (\ref{eq:neww})
     \EndFor
     \State Get new weighted graph $\mathcal{G}^\mathrm{los'}=(\mathcal{V},\mathcal{E}^\mathrm{los'},\mathcal{W}')$
     \State  Solve $\bar{\mathcal{T}}_w^\mathrm{los'}=\argmin_{\{\mathcal{T}_w^\mathrm{los'}\}} \sum_{(v_i,v_j)\in \mathcal{E}^{T}} -w'_{i,j}$ by standard MST algorithm: $\bar{\mathcal{T}}_w^\mathrm{los'}\gets$ MST($\mathcal{G}^\mathrm{los'}$)
     \State  \Return{$\mathbf{u}^* =\argmin_\mathbf{u} \sum_{i=1}^{N}||\mathbf{u}_i-\mathbf{\Hat{u}}_i||^2 $} where $\mathbf{u}\in \mathcal{B}^\mathrm{s}(\mathbf{x})\bigcap\mathcal{B}^\mathrm{obs}(\mathbf{x},\mathbf{x}^\mathrm{obs})\bigcap \mathcal{B}^\mathrm{los}(\mathbf{x},\mathcal{C}_\mathrm{obs},\bar{\mathcal{T}}_w^\mathrm{los'}),||\mathbf{u}_i|| \leq u_\mathrm{max}, \forall i=1,\ldots,N $ \label{alg:line:find_u_star}
     \EndFor
    \EndFunction
    \end{algorithmic}
    % }%
\end{algorithm}
\setlength{\textfloatsep}{0pt}
}

\subsection{Theoretical Analysis}\label{sec:theory_analysis}
In this section, we provide discussions on the validity and feasibility of the proposed LOS-CBC in Lemma~\ref{loscbcdefinition}, the feasibility of the problem (\ref{eq:rawobj}), and how the global and subgroup LOS connectivity is guaranteed over time. 
% Due to the page limit, 
All detailed proofs can be found in the section~\ref{app:sec:valid_cbf_proof} and section~\ref{app:LOS_overtime}, and here we only summarize the main results. 
% \subsubsection{Analysis on LOS-CBC}

\begin{lemma}\label{valid}
Function $h_{i,j,o}^\mathrm{los}(\mathbf{x},\mathbf{x}^\mathrm{obs})$ in (\ref{los1}) is a valid CBF and the admissible control space constrained by~(\ref{eq:los_cbc_definition}) is always non-empty. 
\end{lemma}

With Lemma~\ref{valid}, assuming the robotic team is initially LOS connected, then Lemma~\ref{loscbcdefinition} defines a non-empty (feasible) control space that enforces robots on staying LOS connected.

% \subsubsection{Composition of valid CBFs and feasibility discussions of QP (\ref{eq:rawobj})}
\noindent
\textit{Composition of valid CBFs and feasibility discussions of QP (\ref{eq:rawobj}):}
In this paper, the composition of different valid control barrier functions can be defined as:

\vspace{-0.5cm}
% \begin{footnotesize}
\begin{align}
    h_{i,j,o}^\mathrm{sys} &= h_{i,j,o}^\mathrm{los} \land h_{i,j}^\mathrm{c}\land h_{i,j}^\mathrm{s}\land h_{i,o}^\mathrm{obs} \nonumber\\
    &= \min\{{\min\{{h_{i,j,o}^\mathrm{los},h_{i,j}^\mathrm{c}}\},\min\{{h_{i,j}^\mathrm{s},h_{i,o}^\mathrm{obs}}\}}\}\label{cbf_composition}
\end{align}
% \end{footnotesize}
The composition of CBFs is studied in \cite{capelli2020connectivity,egerstedt2018robot,glotfelter2017nonsmooth} and feasibility analysis under bounded control inputs with mitigation strategies are given in \cite{xiao2022sufficient}. Readers are referred to our appendix in Section~\ref{app:sec:feasible_qp}
% \cite{yupeng}
% \cite{glotfelter2017nonsmooth, xiao2022sufficient}
for detailed discussion. In summary, if the team of robots satisfy the safety and LOS connectivity condition initially, then the presented QP problem(\ref{eq:rawobj}) can always be made feasible that ensures LOS connectivity requirement and safety at all times, e.g. by robots decelerating to zero velocities so that safety and current LOS connectivity graph is preserved even in extreme cases similar to \cite{wang2017safety}.

%\begin{theorem}
%\textbf{Existence of LOS-CBC} Assuming all pairwise robots are initially LOS connected, at $t=0$ i.e. %(\ref{eq:h_conn}) and (\ref{los1}) hold true, then the LOS-CBC defined in %equ.~(\ref{eq:los_cbc_definition}) is guarantee to exist.
%\end{theorem}
%\begin{proof}
%Adding additional CBF can be considered as the composition of CBF:
%\begin{equation}
 %   h^\mathrm{los\land c} = h^\mathrm{los} \land h^\mathrm{c} = \min{(h^\mathrm{los},h^\mathrm{c})}
%\end{equation}
%It was proved in \cite{egerstedt2018robot} and \cite{glotfelter2017nonsmooth} that if the smallest of %$h^\mathrm{los}$ and $h^\mathrm{c}$ is positive then both of them are positive. Then combine the %Theorem~\ref{feasibilitytherom}, there must exist a feasible control $u$ to enforce robot team stay LOS %connected.
%\end{proof}
%\begin{itemize}
%%\item Each node starts as a fragment by itself
%\item Each fragment iteratively connects with MWOE fragment
%\end{itemize}

% \subsubsection{Ensure Global and Subgroup LOS Connectivity over Time} 
% Then we will give theoretical analysis of our Algorithm~\ref{alg:Dynamic} in terms of fulfilling the required global and subgroup LOS connectivity with the computed MLCCST $\bar{\mathcal{T}}_w^\mathrm{los'}$ to preserve. The edges in $\bar{\mathcal{T}}_w^\mathrm{los'}$ are least likely to be broken or require least amount of revision from the nominal task-related controllers $\mathbf{\hat{u}}$.
% \vspace{-5pt}
\begin{proposition}\label{guarantee}
Assume $ \mathcal{G}^\mathrm{los}$ is initially both global and subgroup LOS connected. By 
% assigning the weights in (\ref{WLOSdefine}) and 
following the process 
% of 
% constructing MLCCST $\bar{\mathcal{T}}_w^\mathrm{los'}\subseteq \mathcal{G}^\mathrm{los'}$ 
in Algorithm~\ref{alg:Dynamic} at each time step, it is guaranteed that the resulting communication graph $\mathcal{G}^\mathrm{los}$ in the next time step
% 1) MLCCST $\bar{\mathcal{T}}_w^\mathrm{los'}$ is the spanning sub-graph of the LOS connectivity graph (i.e.$\bar{\mathcal{T}}_w^\mathrm{los'} \subseteq \mathcal{G}^\mathrm{los}$), and 2) the MLCCST $\bar{\mathcal{T}}_w^\mathrm{los'}$ 
is always global and subgroup LOS connected (See proof in Section~\ref{app:LOS_overtime}).
\end{proposition}
\vspace{-0.7cm}
\section{Result}
\begin{figure*}[!htbp]
\centering  
\begin{subfigure}{0.27\textwidth} %0.27
\label{Fig1(a)}
\includegraphics[width=0.8\textwidth,height=0.6\textwidth]{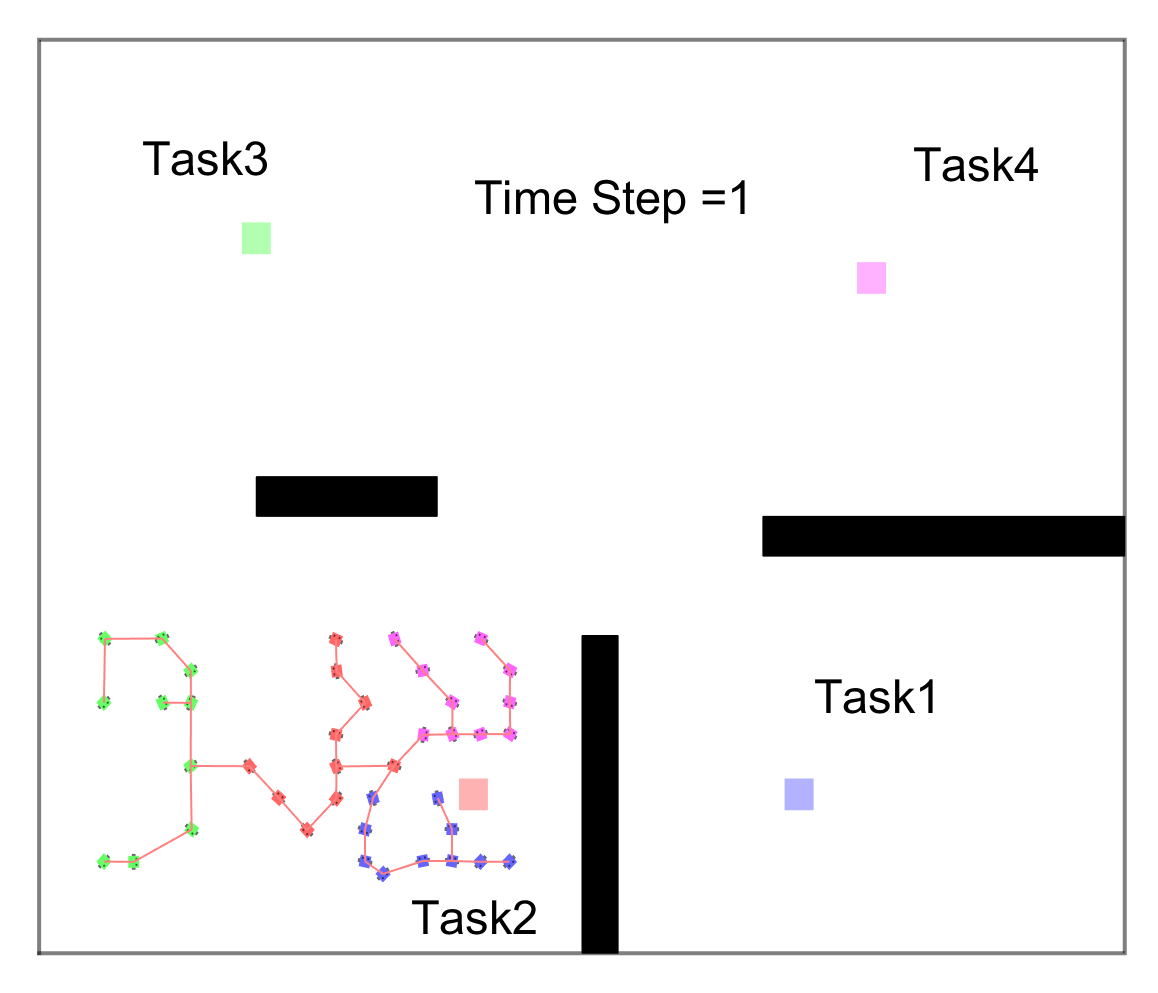}
\caption{t = 1 (MCCST \cite{luo2020behavior})}
\end{subfigure}
\begin{subfigure}{0.27\textwidth}
\label{Fig1(b)}
\includegraphics[width=0.8\textwidth,height=0.6\textwidth]{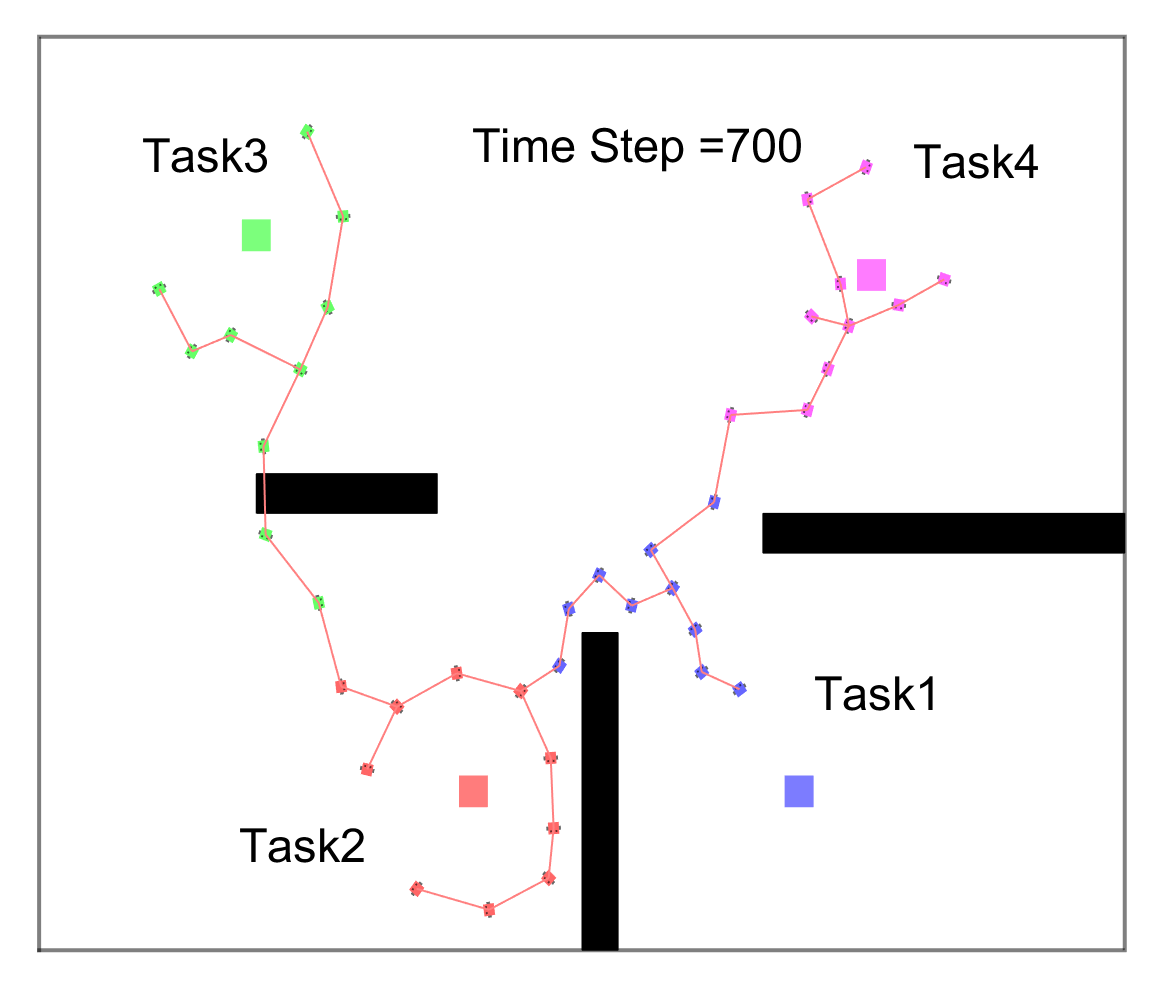}
\caption{t=700 (MCCST \cite{luo2020behavior})}
\end{subfigure}
\begin{subfigure}{0.27\textwidth}
\label{Fig1(c)}
\includegraphics[width=0.8\textwidth,height=0.6\textwidth]{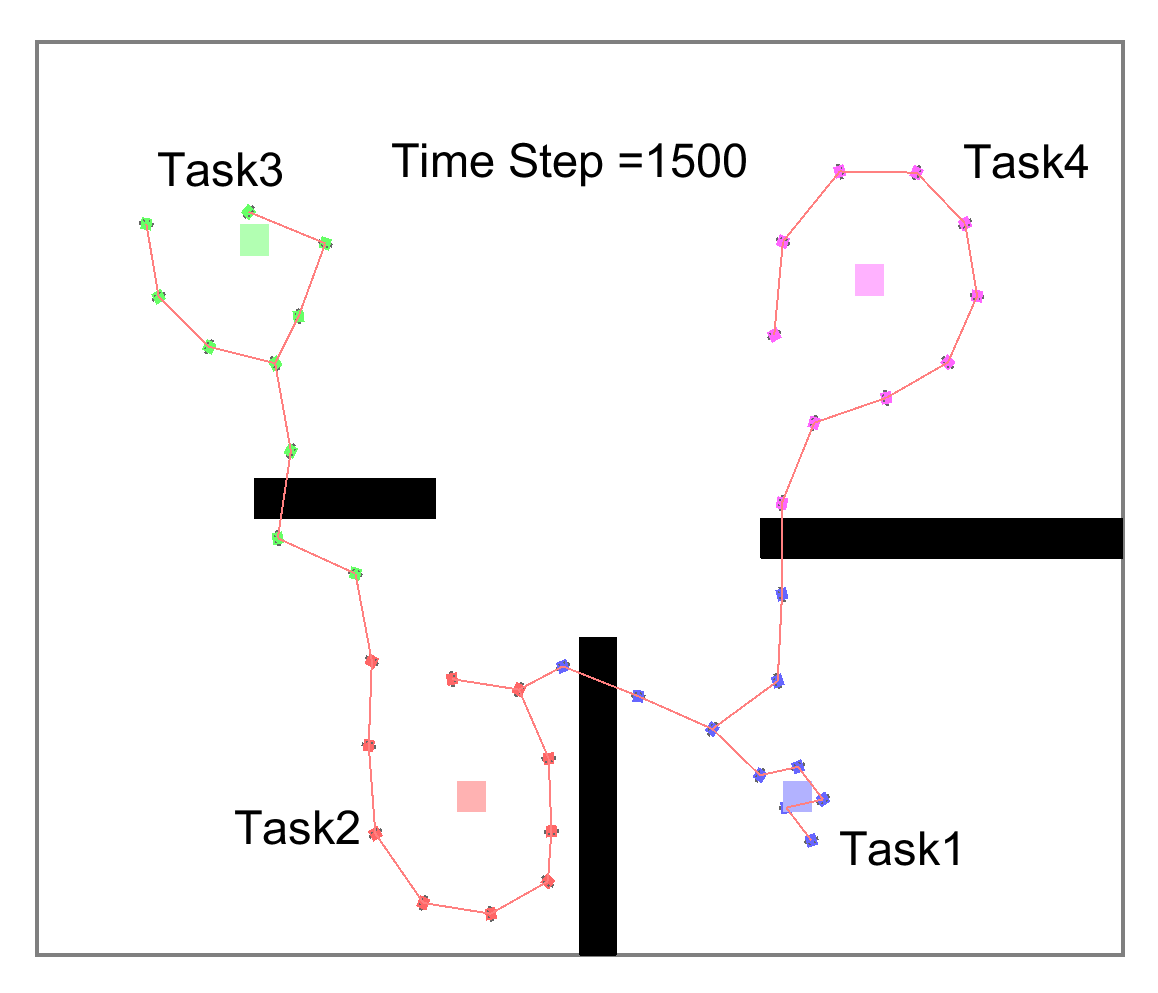}
\caption{t = 1500 (MCCST \cite{luo2020behavior}, Converged)}
\end{subfigure}
\begin{subfigure}{0.27\textwidth}
\label{Fig1(d)}
\includegraphics[width=0.8\textwidth,height=0.6\textwidth]{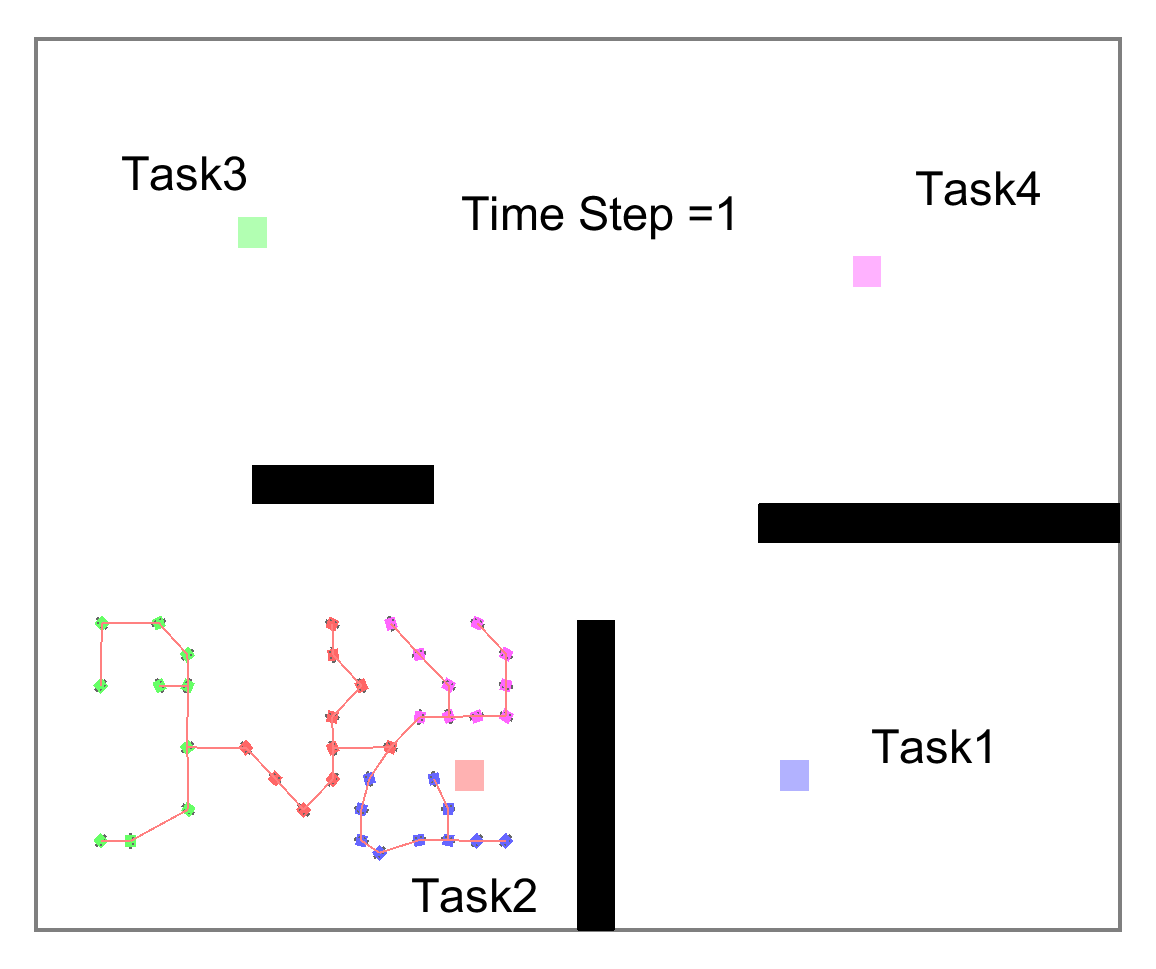}
\caption{t = 1 (MLCCST)}
\end{subfigure}
\begin{subfigure}{0.27\textwidth}
\label{Fig1(e)}
\includegraphics[width=0.8\textwidth,height=0.6\textwidth]{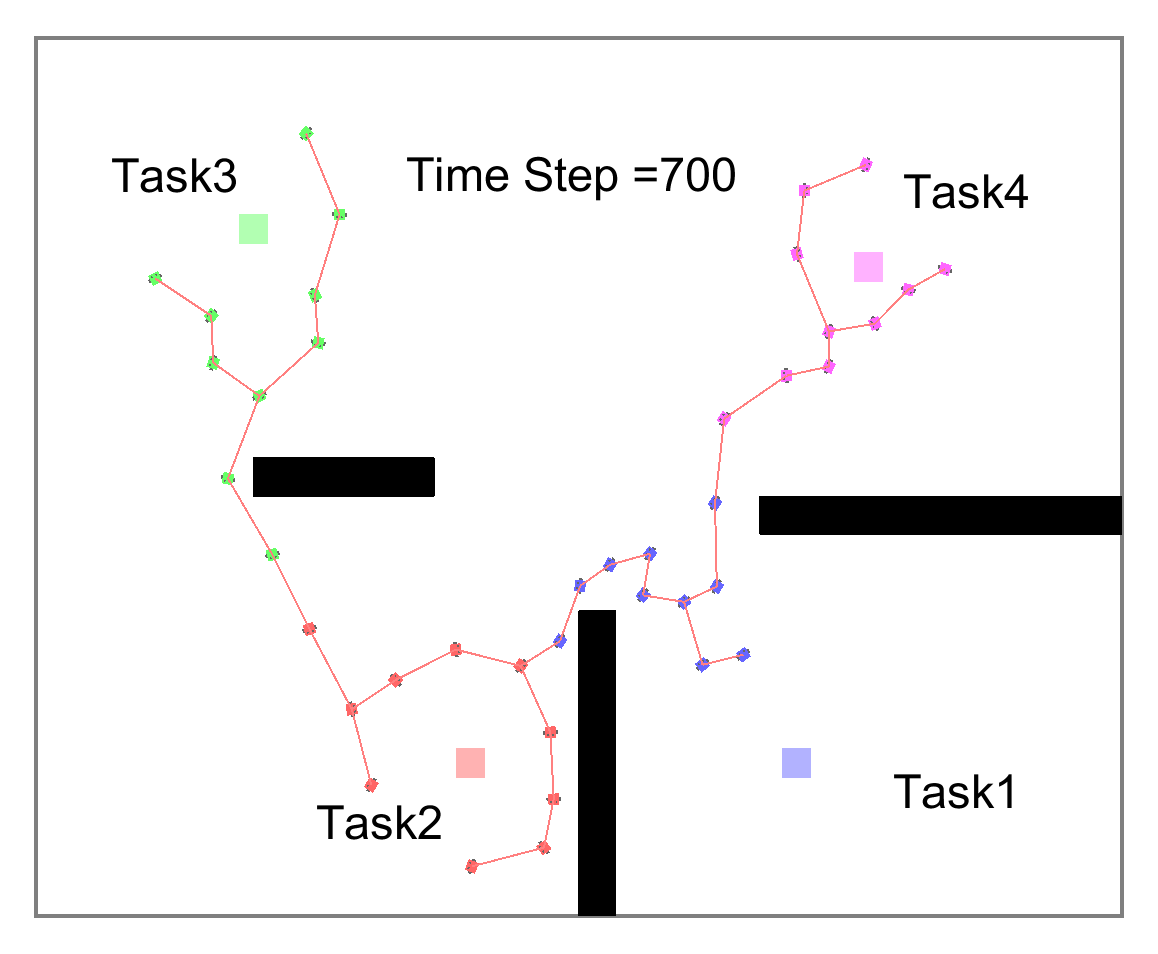}
\caption{t=700 (MLCCST )}
\end{subfigure}
\begin{subfigure}{0.27\textwidth}
\label{Fig1(f)}
\includegraphics[width=0.8\textwidth,height=0.6\textwidth]{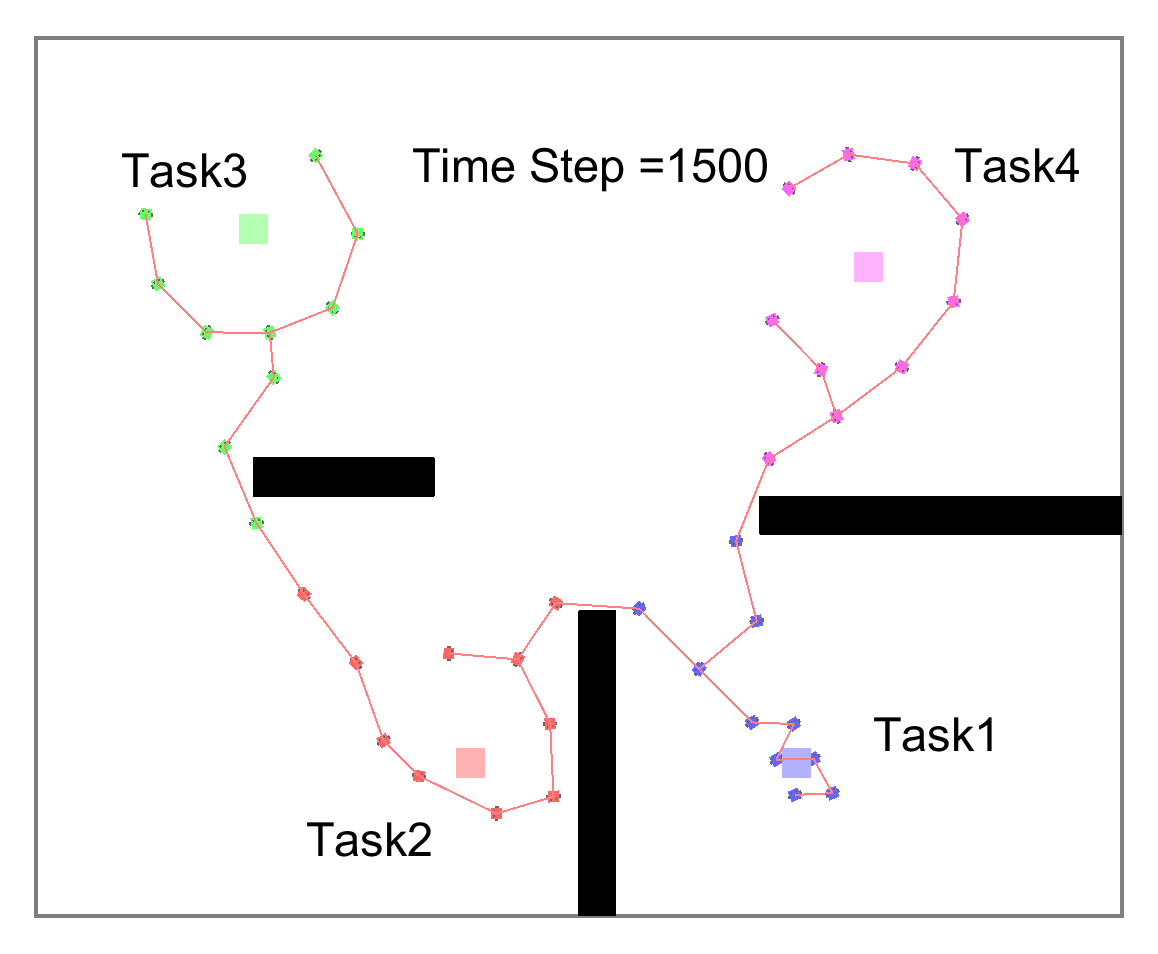}
\caption{t=1500 (MLCCST, Converged)}
% \vspace{-10pt}
\end{subfigure}

\caption{{\footnotesize Comparison of our proposed MLCCST (second row) with MCCST \cite{luo2020behavior} (first row). Our proposed MLCCST with the Line-of-Sight Connectivity Barrier Certificate can ensure the LOS connectivity over time.
However, the MCCST can't ensure the LOS connectivity (The communication edges are cutoff by obstacles in Figure (b) and (c)). t is time step.
The red lines in this figure denote the currently active line of sight connectivity graph. The black boxes represent the obstacles. The robot team is divided into $M=4$ subgroup with different colors and is tasked with 4 parallel behaviors. In the figures, robots in blue subgroup 1 execute biased rendezvous behaviors towards the blue task site 1, while robots in red subgroup 2, green subgroup 3, and magenta subgroup 4 perform circle formation behaviors around the red task site 2, green task site 3 and magenta task site 4 respectively.
% Compared to the MCCST \cite{luo2020behavior} (a), (b) and (c), our proposed MLCCST approach (d), (e) and (f) could ensure the LOS connectivity (the black obstacles do not cut off the connectivity edges) over time.
}}
\label{Fig1}
% \vspace{-10pt}
\end{figure*}

\begin{figure*}[!htbp]
\centering  
\begin{subfigure}{0.3\textwidth}
\label{Fig2(a)}
\includegraphics[width=0.8\textwidth,height=0.6\textwidth]{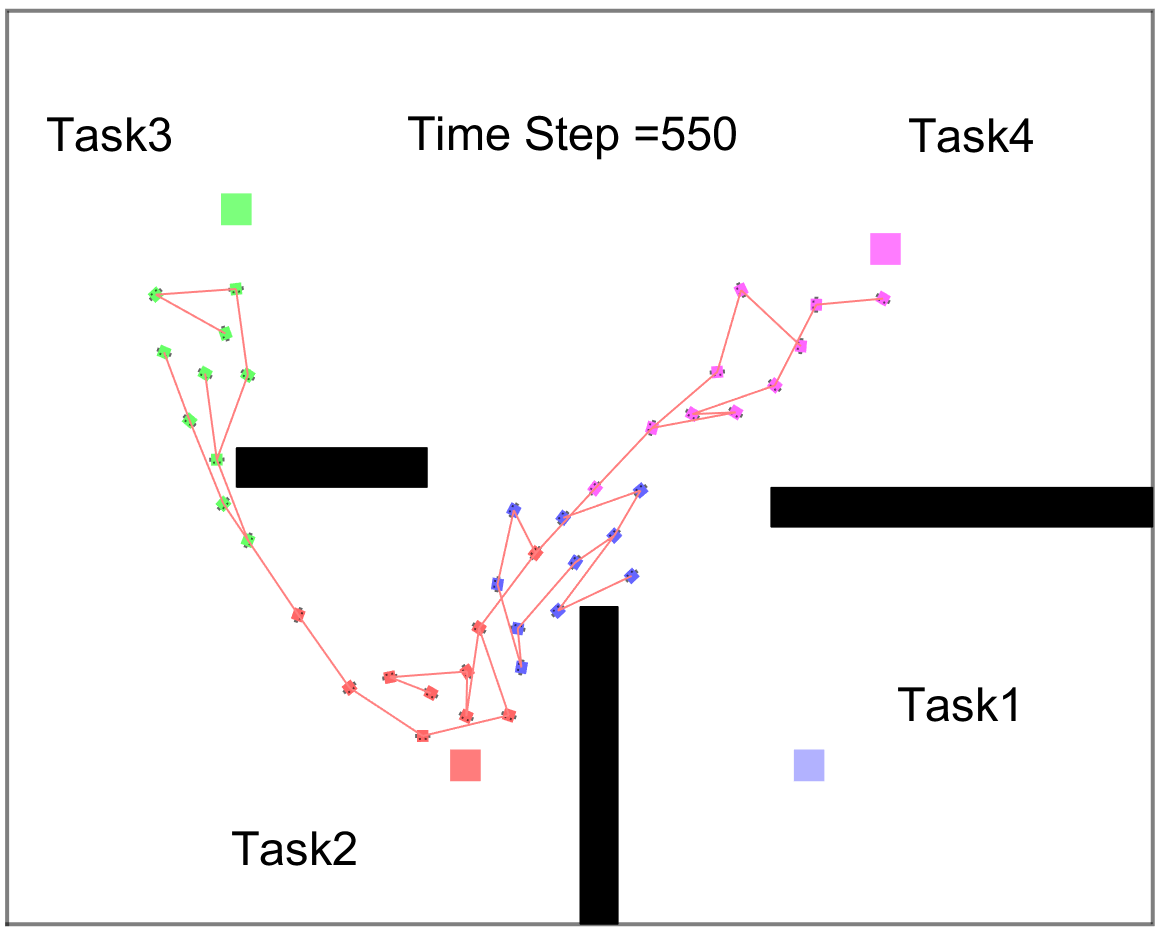}
\caption{t=550 (Fixed MLCCST)}
\end{subfigure}
\begin{subfigure}{0.3\textwidth}
\label{Fig2(b)}
\includegraphics[width=0.8\textwidth, height=0.6\textwidth ]{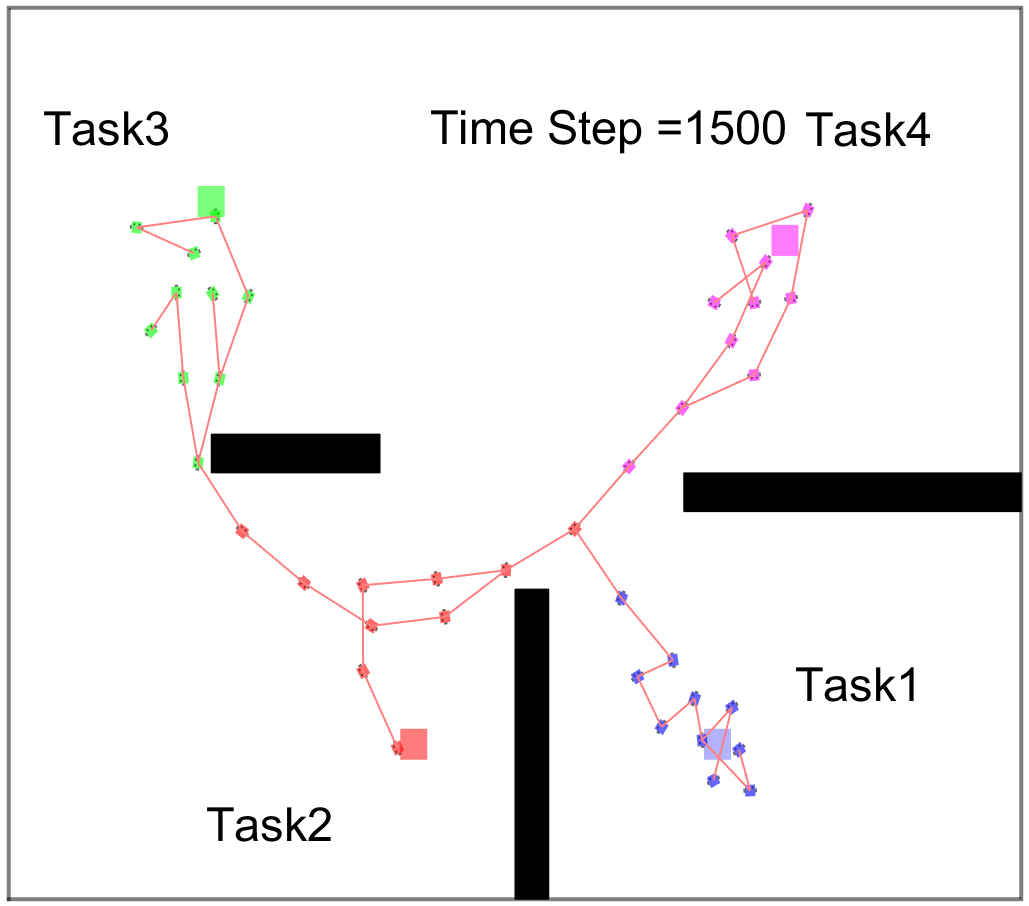}
\caption{t=1500 (Fixed MLCCST, Converged)}
\end{subfigure}
\caption{Performance of Fixed MLCCST (our proposed method without dynamically updated LOS connectivity graph in runtime). With the fixed initial MLCCST, the team of robots could ensure the LOS connectivity, but the task performance will be perturbed greatly.
% (Subgroup of robots fail to perform their original tasks)
}
\label{Fig2}
\vspace{-10pt}
\end{figure*}
\setlength{\textfloatsep}{0pt}
\begin{figure*}[!htbp]
\centering  
\begin{subfigure}[t]{0.24\textwidth}
\label{Fig3(a)}
\includegraphics[width=\textwidth]
{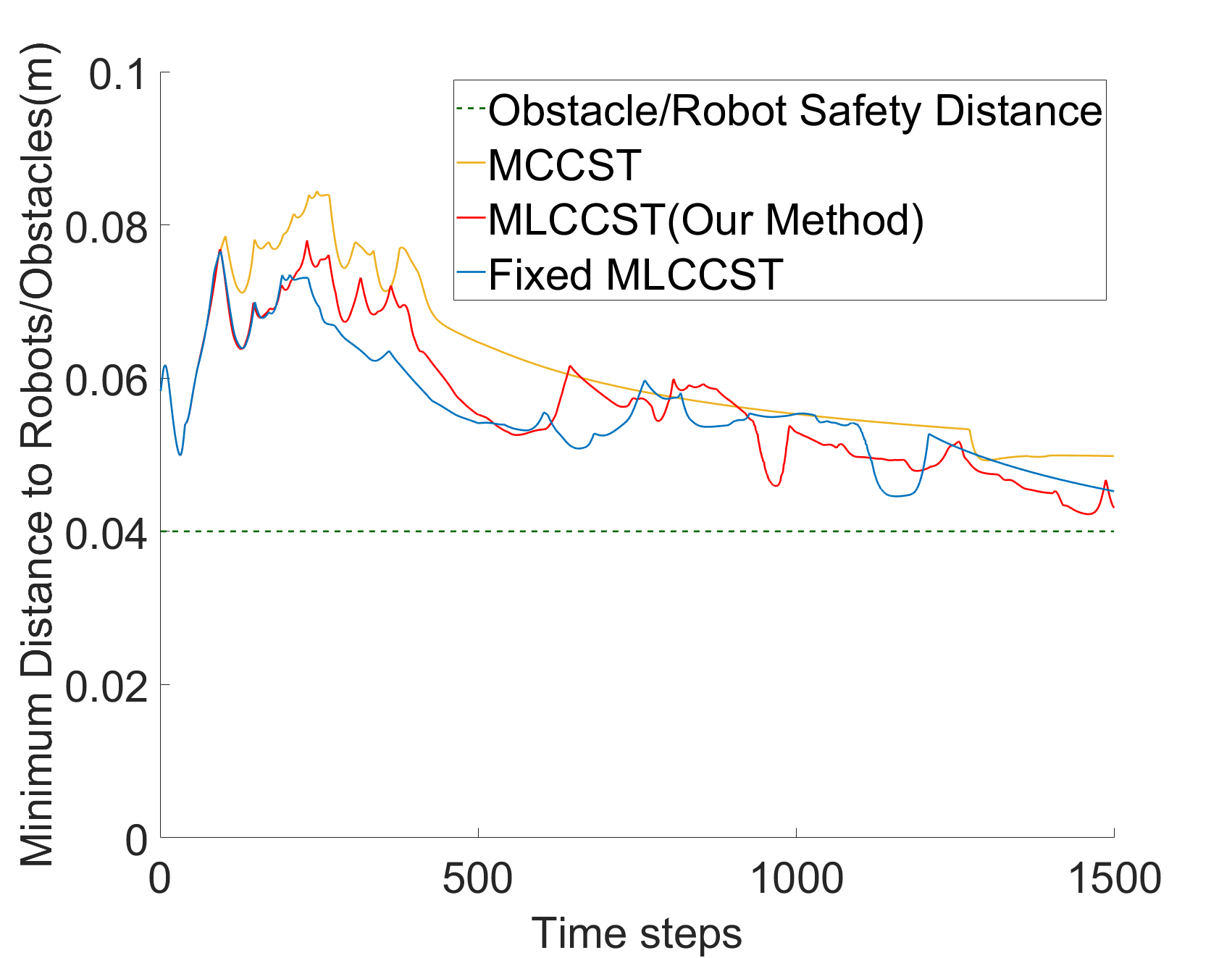}
\caption{$D_{min}$ to robots/obstacles}
\end{subfigure}
\begin{subfigure}[t]{0.24\textwidth}
\label{Fig3(b)}
\includegraphics[width=\textwidth]{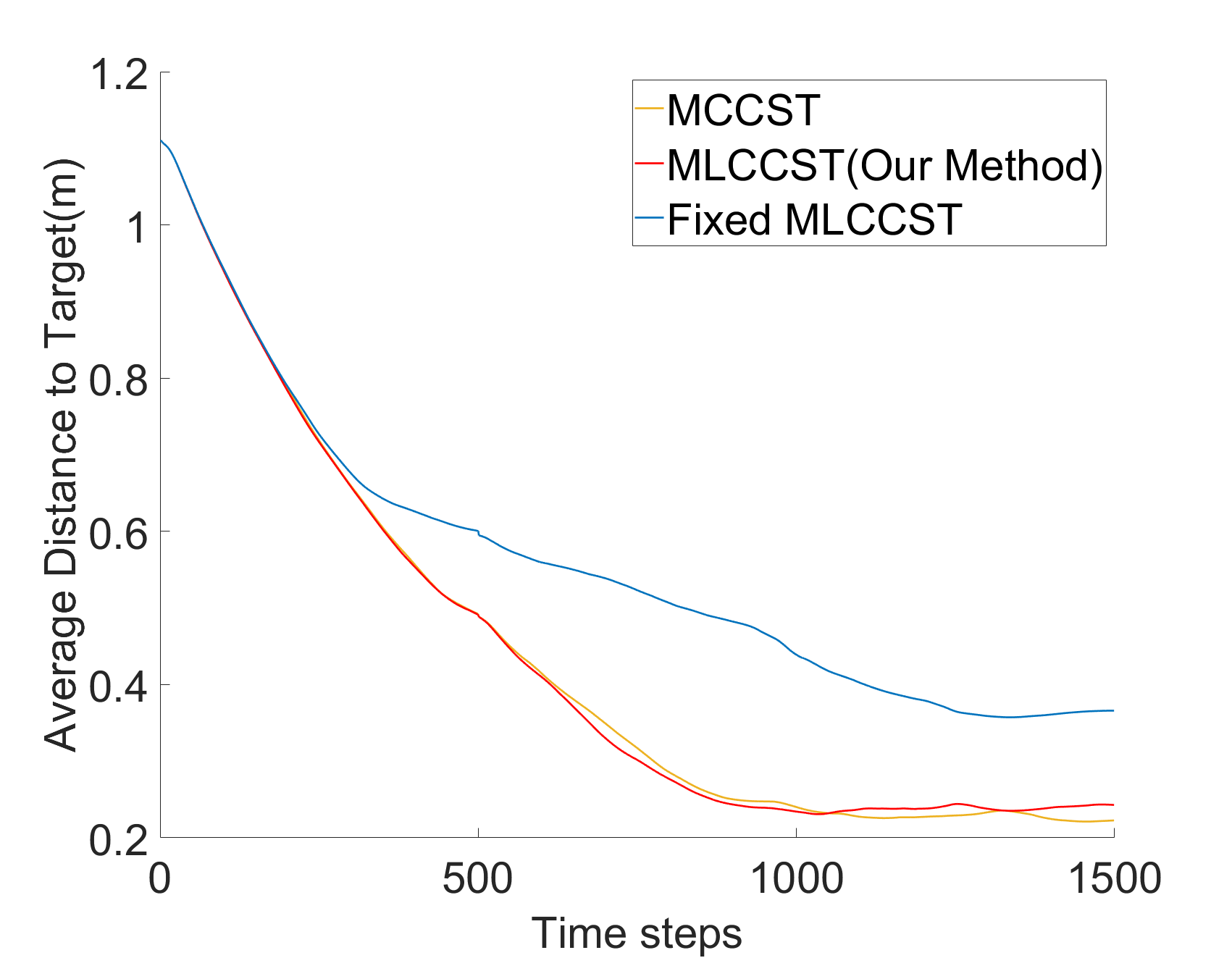}
\caption{$D_{ave}$ to target region}
\end{subfigure}
\begin{subfigure}[t]{0.24\textwidth}
\label{Fig3(c)}
\includegraphics[width=\textwidth]{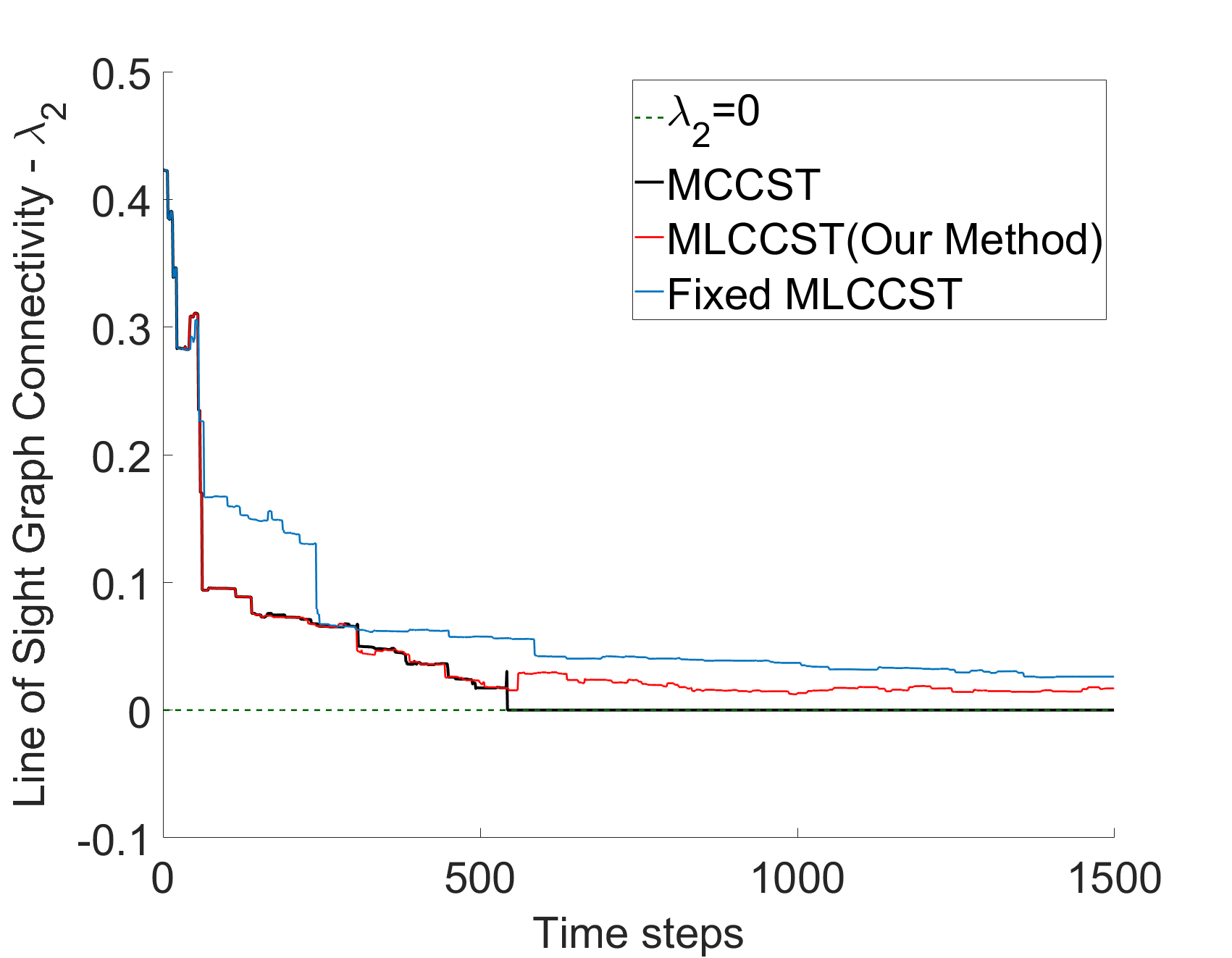}
\caption{Algebraic LOS connectivity}
\end{subfigure}
\begin{subfigure}[t]{0.24\textwidth}
\label{Fig3(d)}
\includegraphics[width=\textwidth]{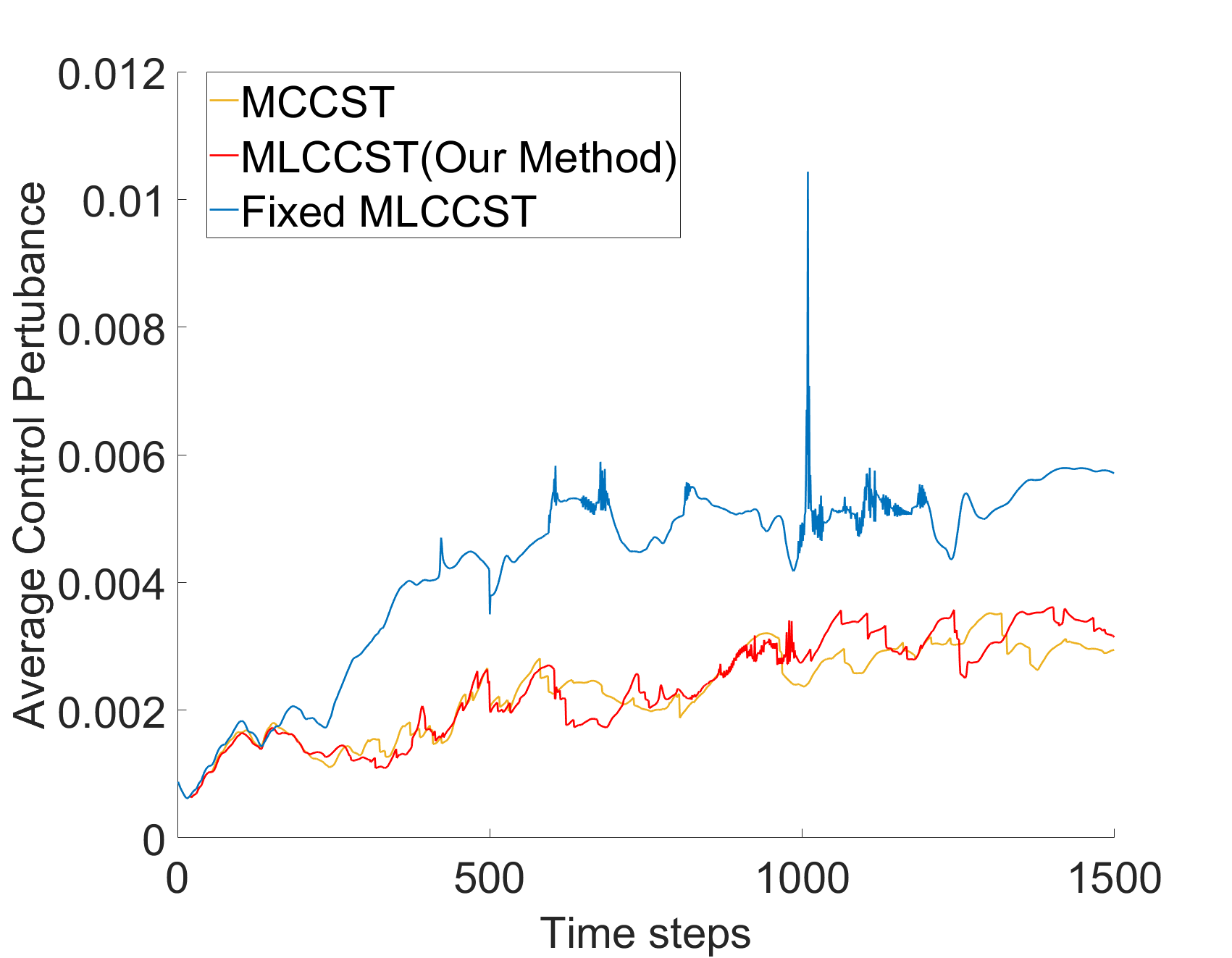}
\caption{Average control perturbation}
\end{subfigure}
\caption{ {\footnotesize Performance comparison of the proposed MLCCST and two baseline methods on the simulation example demonstrated in Figure~\ref{Fig1} and Figure~\ref{Fig2} with respect to four selected metrics: (a) Minimum distance to robots/obstacles (verify safety constraint satisfaction and degree of conservativeness of the system) (safety distance is 0.04m), (b) Average distance between robots to tasked region (indicate the overall task efficiency), (c) Average algebraic LOS connectivity (indicate whether the LOS graph is LOS connected ($\lambda_2 >0$) or not ($\lambda_2 =0$), $\lambda_2$ is the second smallest eigenvalue of the Line-of-Sight laplacian matrix calculated from the Line-of-Sight adjacency matrix. The elements in the Line-of-Sight adjacency matrix indicate whether the pairwise robots are LOS connected), (d) Average Control perturbation (computed by $\frac{1}{N}\sum_{i=1}^N||
\mathbf{u}_i-\Hat{\mathbf{u}}_i||^{2}$ to measure the accumulated deviation from nominal controllers).
} }
%computed by $\frac{1}{N}\sum_{i=1}^N||u_i^*-\Hat{u}_i|
%|^2$}.
\label{Fig3}
\vspace{-10pt}
\end{figure*}

\begin{figure*}[!htbp]
\centering  
\begin{subfigure}[t]{0.19\textwidth}
\label{Fig4(e)}
\includegraphics[width=\textwidth]{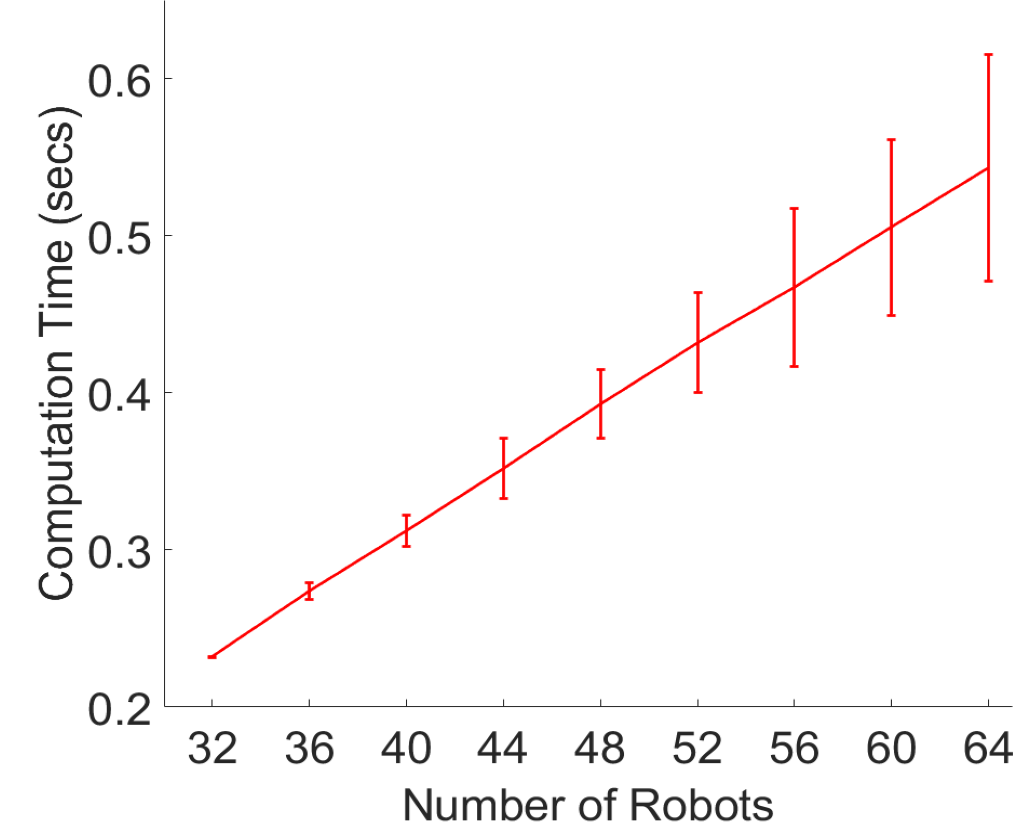}
\caption{Average computation time}
\end{subfigure}
\begin{subfigure}[t]{0.19\textwidth}
\label{Fig4(a)}
\includegraphics[width=\textwidth]{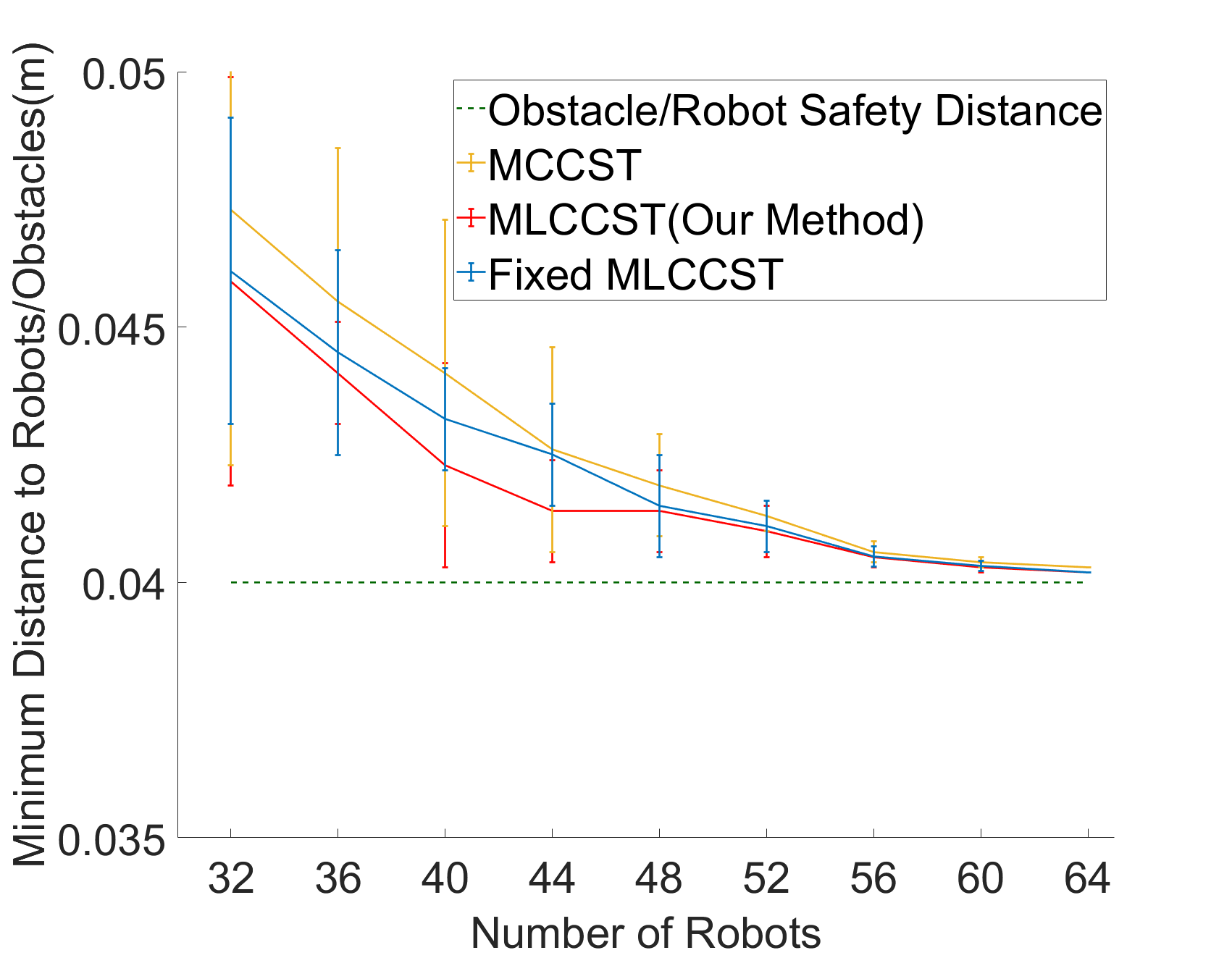}
\caption{robots/obstacles $D_{min}$}
\end{subfigure}
\begin{subfigure}[t]{0.19\textwidth}
\label{Fig4(c)}
\includegraphics[width=\textwidth]{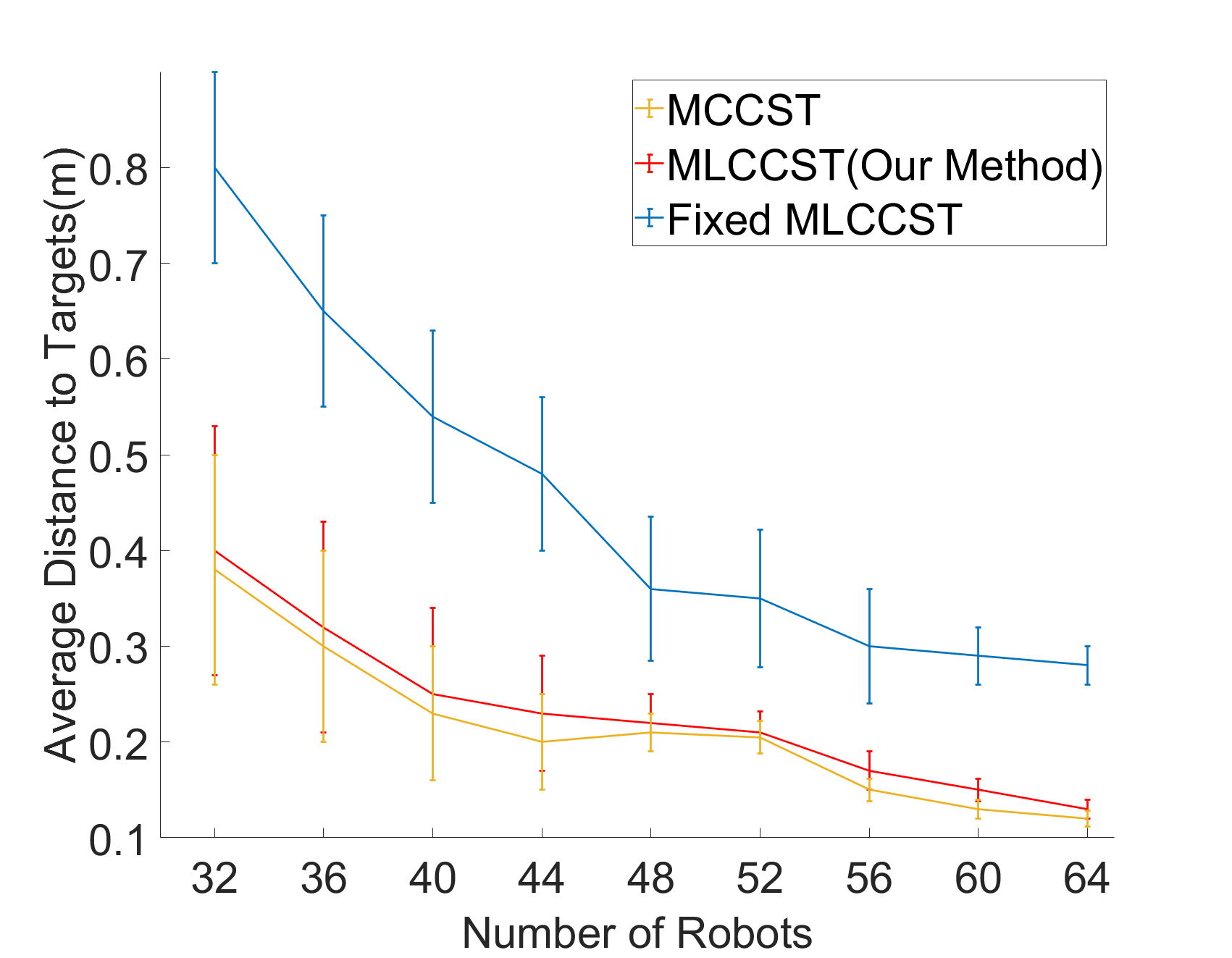}
\caption{$D_{ave}$ to target region}
\end{subfigure}
\begin{subfigure}[t]{0.19\textwidth}
\label{Fig4(b)}
\includegraphics[width=\textwidth]{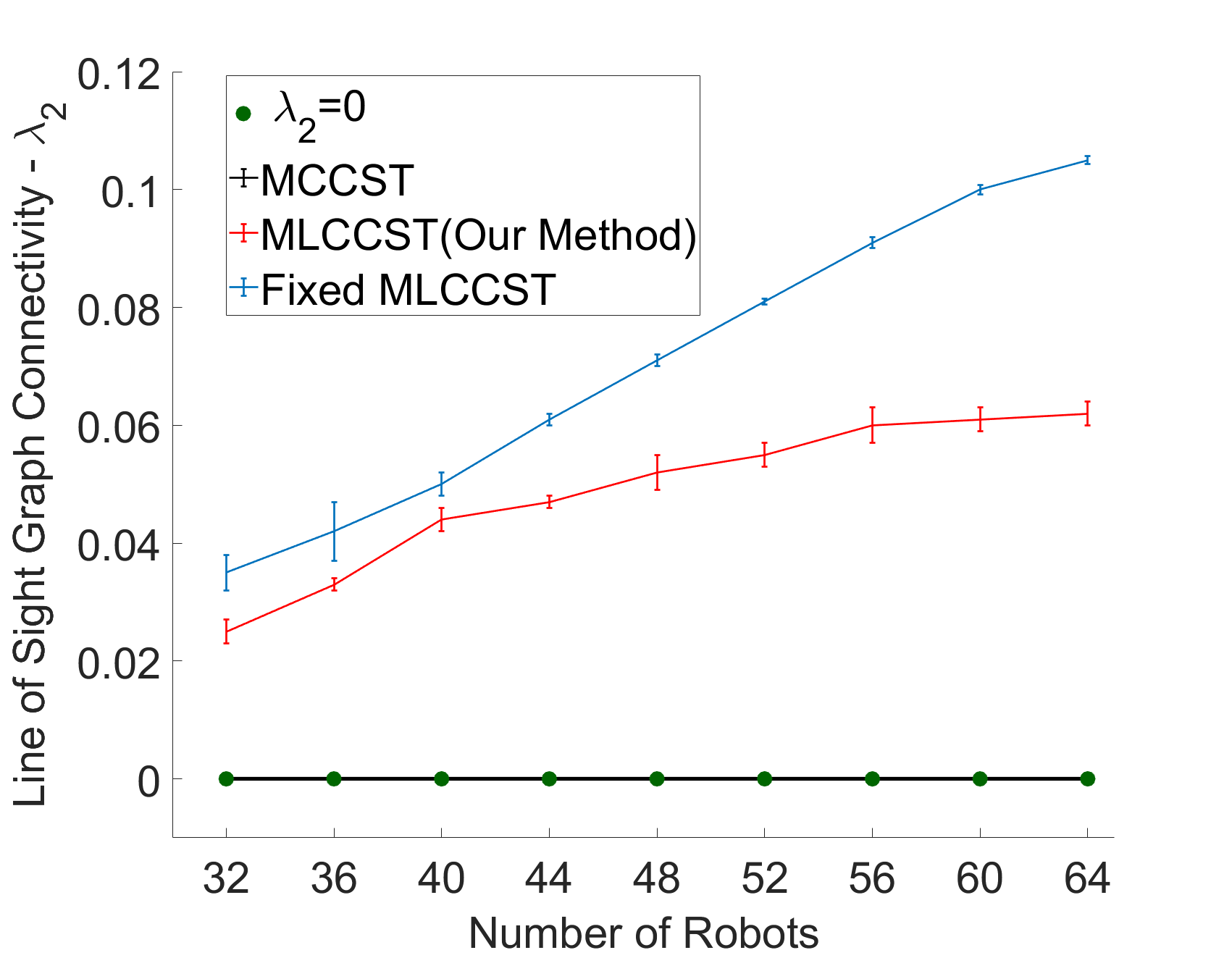}
\caption{Algebraic LOS connectivity}
\end{subfigure}
\begin{subfigure}[t]{0.19\textwidth}
\label{Fig4(d)}
\includegraphics[width=\textwidth]{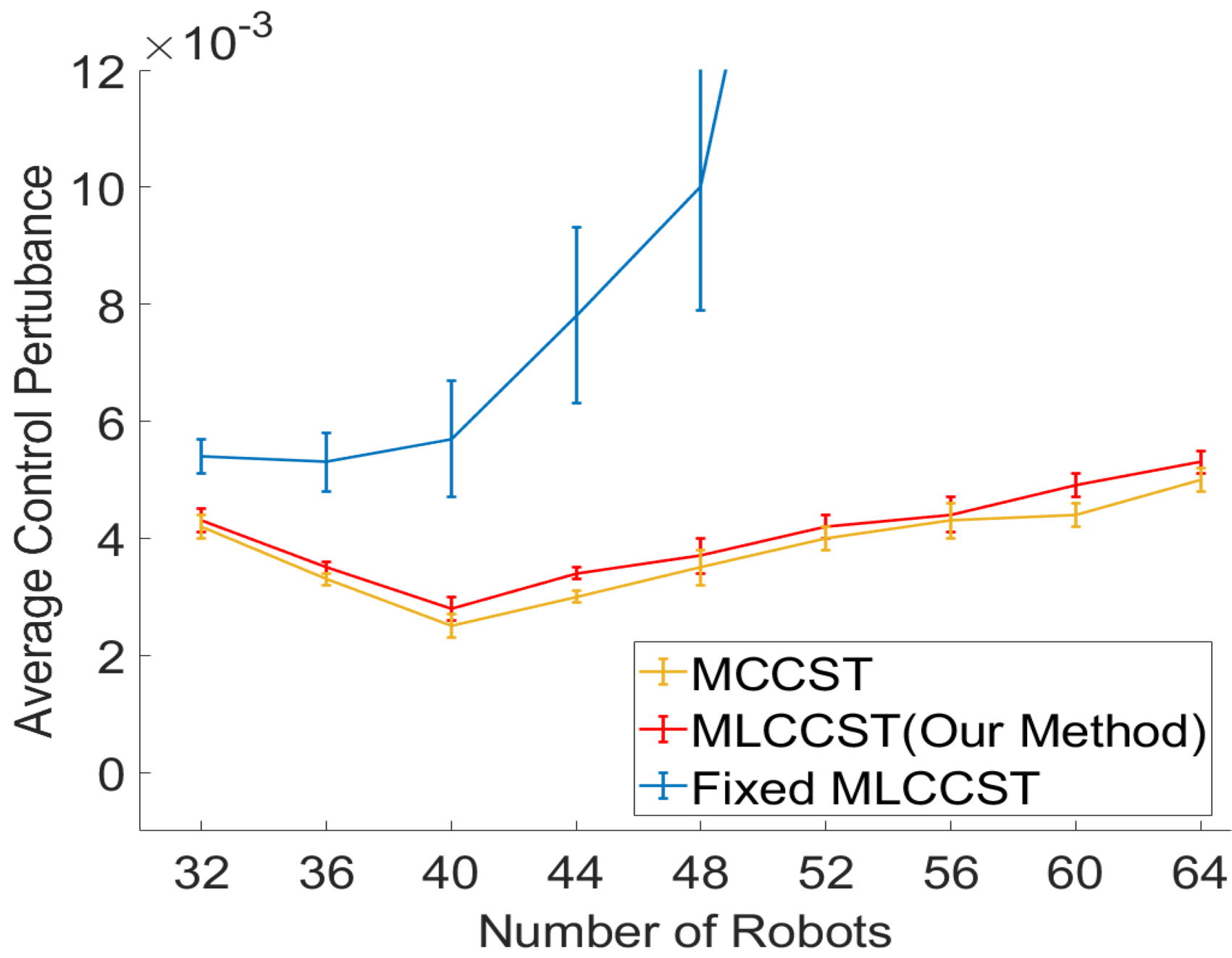}
\caption{Average control perturbation}
\end{subfigure}

\caption{Quantitative results of performance on different sizes of robot team with respect to five selected metrics. For all figures, the error bar shows the standard deviation.}
\label{Fig4}
\vspace{-10pt}
\end{figure*}
\subsection{Simulation Example}
The first set of experiments performed on a team of $N=40$ with unicycle dynamics are shown in Figure~\ref{Fig1} and Figure~\ref{Fig2}. 
% To solve the MVEE optimization problem, we adopt the solution from \cite{Anye}. 
We apply the minimally revised controllers from (\ref{eq:rawobj}) to the robots with unicycle dynamics using kinematics mapping from \cite{wang2017safety}. 
%\wl{we should use (b) 
%Most of the behavior configurations have been accomplished in Figure~\ref{Fig1}(f).
To demonstrate 1) the effectiveness of the proposed LOS-CBC, and 2) the importance of updating the LOS connectivity graph dynamically over time, two baseline methods are implemented for performance comparison: Minimum Connectivity
Constraint Spanning Tree (MCCST) \cite{luo2020behavior} which only considers the regular connectivity constraints, and Fixed MLCCST, which preserves the initial MLCCST computed from the first time step with our method but without dynamically updating over time.
% which does not dynamically update the LOS connectivity graph. 
For MCCST, without considering the line-of-sight connectivity constraints, the robots easily lose inter-robot LOS communications due to obstacle occlusions. 
% Compared to Figure~\ref{Fig1}(a), our MLCCST algorithm enforces provably minimum LOS connectivity graph which is them same graph in MCCST that is least restrictive to robots. Note in \cite{luo2020behavior}, we have already proved that this connectivity graph is least restrictive to robots. 
In Figure~\ref{Fig2}, it is observed that without dynamically updating the LOS connectivity graph as robots move, the task performance is significantly affected. 
% Since the fixed connectivity graph will introduce unnecessary connectivity constraints, it will even trap the robot team in a deadlock. 

Figure~\ref{Fig3}(a) shows that all three algorithms satisfy the safety requirement with no collision happening. Figure~\ref{Fig3}(b) indicates that our proposed MLCCST method achieves almost the same high task efficiency as MCCST does. Note that MCCST performs slightly better than our proposed method as MCCST does not have Line-of-Sight connectivity constraints and therefore robot motion is less restrictive.
Without dynamically updating the connectivity graph, the performance of Fixed MLCCST shows that the fixed connectivity topology graph will limit the task performance significantly. Figure~\ref{Fig3}(c) shows that both our proposed MLCCST and Fixed MLCCST ensure the Line-of-Sight connectivity during the task, but robots using MCCST becomes disconnected in terms of Line-of-Sight ($\lambda_2=0$). Figure~\ref{Fig3}(d) shows that Fixed MLCCST introduced the most average control perturbation,
while our MLCCST method achieves the least average control perturbation which reflects the task efficiency due to our minimally constraining design. In summary, based on the four selected evaluation metrics, our proposed MLCCST method performs the best among all three algorithms.
\subsection{Quantitative Results}
To validate the computation efficiency and scalability of our algorithm, we run experiments with up to 64 robots and 4 parallel behaviors. We conducted 10 trials for each batch of robots with varying sizes. 
% The performance comparison of the proposed MLCCST and baseline methods are shown in Figure~\ref{Fig4}.
Figure~\ref{Fig4}(a) shows the average computation time with our MLCCST and indicates its good computation efficiency for real-time application. Figure~\ref{Fig4}(b)-(c) demonstrate that collision-free motion is achieved for all three methods and our MLCCST method achieves good task efficiency in terms of reducing distance to target region. 
% while our proposed method is the least conservative and the task efficiency is significantly higher when the robot team size is smaller than 48. 
Figure~\ref{Fig4}(d) shows that both Fixed MLCCST and our proposed MLCCST maintain the satisfying LOS connectivity,
% , verifying the effectiveness of our proposed Line-of-Sight Connectivity Barrier Certificate,
while MCCST fails to preserve it ($\lambda_{2} = 0$) as it does not consider LOS connectivity. Figure~\ref{Fig4}(e) indicate that both MCCST and our proposed MLCCST achieve 
% shorter average distance to targets and 
comparable average control perturbance much less than Fixed MLCCST, inferring more flexibility on the robot motion by our MLCCST despite the LOS connectivity constraints.
% The main reason that Fixed MLCCST does not perform as well as the other two is that, the fixed connectivity topology traps the robot team in the deadlock situation and prevents them from pursuing assigned tasks.
% The performance of MCCST is slightly better than our proposed method due to the ignorance of the line-of-sight connectivity constraint.
% Therefore, there will be fewer agents maintaining connectivity and more agents completing assigned tasks.
\vspace{-0.3cm}
\section{Conclusion} 
\label{sec:conclusion}
In this work, we propose a novel notion of Line-of-sight Connectivity Barrier Certificate (LOS-CBC) to define admissible control space that guarantees LOS connectivity between pairwise robots over time. By combining LOS-CBC and a graph theoretic approach, we formulate the global and subgroup LOS-aware multi-robot coordination problem as a bilevel optimization, and the Minimum LOS Connectivity Constraint Spanning Tree (MLCCST) algorithm is proposed to realize the minimal deviation control subject to safety and optimized LOS connectivity constraints. Theoretical analysis and discussions are provided to justify the performance of our proposed MLCCST algorithm. 
Future work includes 
% to \yy{investigate alternative weight assignment algorithms that can more accurately reflect the level of constraint violations on each edge.} and
% wenhao: I took out this since it's a bit too detailed to discuss here, and we will address this in future journal version
to fully decentralize the MLCCST algorithm so that it could run on distributed multi-robot systems.

\bibliographystyle{IEEEtran}
\bibliography{ref}

\onecolumn

\newpage
\section{Appendices}
Equation indexes from (1)-(23) follow the original indexes appearing in the paper and new equations start from (24) in this appendix.

\begin{remark}
In this paper, we adopt the Minimum Volume Enclosing Ellipsoid (MVEE) algorithm \cite{todd2007khachiyan} to approximate the LOS communication edge for each pair-wise robot $\forall (v_i,v_j)\in \mathcal{E}^\mathrm{slos}$ that renders the analytical form of $h_{i,j,o}^\mathrm{los}$ in (\ref{los1}), expressing the occlusion-free condition $h_{i,j,o}^\mathrm{los}\geq 0$. This method is general and can be used to address more challenging environment with uncertainty, e.g. using MVEE to cover the uncertainty elllipsoids on the noisy position measurements of robots when their exact locations are unknown. Although this is beyond the scope of this paper, generalizing our approach to robotic team under positional uncertainties is an important future direction.
% has the potential advantage to solve the LOS connectivity maintenance problem in an uncertain environment, e.g. when the exact locations of robots are described by a random variable 
% the dynamical system has the measurement noise. In this case, the ellipsoidal approximation method can be adopted to cover the state distribution of each robot and don't need to sample points around the midpoint of the communication edge.
In this paper, the particular choice of $2d$ points for constructing MVEE between robots $i,j,\forall (v_i,v_j)\in \mathcal{E}^\mathrm{slos}$ are analytically defined as $\mathcal{P}_{i,j}= \{\mathbf{p}^{1}_{i,j},...,\mathbf{p}^{2d}_{i,j}\}$, where 
$\mathbf{p}^{1}_{i,j}=\mathbf{x}_i,
\mathbf{p}^{2}_{i,j}=\mathbf{x}_j$ are the two vertices on the major principle axis of the MVEE and the other vertices on the non-major principle axis are
% $\mathbf{p}^{p}_{i,j}=\mathbf{p}^{0}_{i,j}+\delta\cdot (-1)^p, p=3,\ldots, 2d$ 
$\mathbf{p}^{2p-1}_{i,j}=\mathbf{p}^{0}_{i,j}+\delta\cdot \mathbf{e}_p,\; \mathbf{p}^{2p}_{i,j}=\mathbf{p}^{0}_{i,j}-\delta\cdot \mathbf{e}_p,\;\forall p=2,\ldots, d$ with $\mathbf{p}^{0}_{i,j} = \frac{\mathbf{x}_i+\mathbf{x}_j}{2}$ and  $\mathbf{e}_p\in\mathbb{R}^d$ as the $p$th unit orthogonal basis of $\mathbb{R}^d$ perpendicular to $\mathbf{x}_i-\mathbf{x}_j$. $\delta\in\mathbb{R}$ is a small pre-defined value that reflects the "thickness" of the ellipsoid.
\end{remark}
% In the main text, we didn't give the analytical form of each point in $\mathcal{P}_{i,j}$, here we give the explicit formulation of each points in set . $\mathbf{p}^{1}_{i,j} = \mathbf{x}_i$, $\mathbf{p}^{2}_{i,j} = \mathbf{x}_j$ and the other points on the non-major principle axis of the MVEE:
% \begin{align}\label{eq:app:mvee}
%   p^{l}_{i,j} = p^{0}_{i,j} \pm A_{k}\delta  \quad \forall l =3,...,2d, \forall k =1,...,d
% \end{align}
% where $A_{k}$ is the unit normal vector of the communication edge in each dimension. In~(\ref{eq:app:mvee}), we sample two points on each non-major principle axis of the MVEE, and the sampled points have the same semi axis length with the middle point $\mathbf{p}^{0}_{i,j} = \frac{\mathbf{x}_i+\mathbf{x}_j}{2}$ as $||\mathbf{p}^{l}_{i,j}-\mathbf{p}^{0}_{i,j}||=\delta,\forall l=3,\ldots,2d $. In summary, by constructing the $\mathcal{P}_{i,j}$ in this way, it is sufficient to determine a MVEE. 

\subsection{Proof of Lemma~\ref{valid}}\label{app:sec:valid_cbf_proof}

\begin{lemma}\label{lem:farkas}
(Variant of Farkas' Lemma \cite{cook2011combinatorial}) Let $\mathbf{A}\in\mathbb{R}^{m\times n}$ and $\mathbf{b}\in\mathbb{R}^{m}$, then exactly one of the following is true for the system $\mathbf{A}\mathbf{x} \leq \mathbf{b}$:
\begin{enumerate}
    \item There is a solution $\mathbf{x}\in\mathbb{R}^n$ such that $\mathbf{A}\mathbf{x} \leq \mathbf{b}$.
    \item There is a solution $\mathbf{y}\geq 0\in\mathbb{R}^m$ such that $\mathbf{y}^\top \mathbf{A} = 0$ and $\mathbf{y}^\top \mathbf{b} < 0$.
\end{enumerate}
\end{lemma}

\begin{customlemma}{5}
Function $h_{i,j,o}^\mathrm{los}(\mathbf{x},\mathbf{x}^\mathrm{obs})$ in (\ref{los1}) is a valid CBF and the admissible control space constrained by~(\ref{eq:los_cbc_definition}) is always non-empty. 
\end{customlemma}
\begin{proof}
Firstly, we demonstrate that the proposed function $h_{i,j,o}^\mathrm{los}(\mathbf{x},\mathbf{x}^\mathrm{obs})$ is a valid control barrier function (CBF) assuming the robotic team is initially global and subgroup LOS connected with $\forall (v_i,v_j)\in \mathcal{E}^\mathrm{slos}$. 
As summarized in \cite{capelli2020connectivity}, from Lemma~\ref{lem:cbf} the condition for a function $h(\mathbf{x})$ to be a valid CBF should satisfy the following three conditions: (a) $h(\mathbf{x})$ is continuously differentiable, (b) the first-order time derivative of $h(\mathbf{x})$ depends explicitly on the control put $\mathbf{u}$ (i.e. $h(\mathbf{x})$ is of relative degree one), and (c) it is possible to find an extended class-$\mathcal{K}$ function $\kappa(\cdot)$ such that $\sup_{\mathbf{u}\in\mathcal{U}}\{\dot{h}(\mathbf{x},\mathbf{u})+\kappa(h(\mathbf{x}))\}\geq 0$ for all $\mathbf{x}$. 

Hence considering our proposed candidate CBF $h_{i,j,o}^\mathrm{los}(\mathbf{x},\mathbf{x}^\mathrm{obs})$ in (\ref{los1}) for $\forall (v_i,v_j)\in \mathcal{E}^\mathrm{slos}$, we have 
%should be continuous differentibale of relative degree one, and it is possible to find an extend class $\kappa$ function $\kappa(h(\mathbf{x}))$ such that equation  $\dot{h}^\mathrm{los}_{i,j,o}(\mathbf{x}, \mathbf{x}^\mathrm{obs}, \mathbf{u})+\gamma h^\mathrm{los}_{i,j,o}(\mathbf{x}, \mathbf{x}^\mathrm{obs}) \geq 0$ is satisfied.
\begin{equation}\label{d}
\begin{split}
    &\frac{\partial{h_{i,j,o}^\mathrm{los}(\mathbf{x},\mathbf{x}^\mathrm{obs})}}{\partial \mathbf{x}} = 2(\mathbf{x}^\mathrm{obs}_{o}-\frac{\mathbf{x}_i+\mathbf{x}_j}{2})^{T}Q_{i,j}\\
    &\frac{d}{dt}h_{i,j,o}^\mathrm{los}(\mathbf{x},\mathbf{x}^\mathrm{obs}) = -(\mathbf{x}^\mathrm{obs}_{o}-\frac{\mathbf{x}_i+\mathbf{x}_j}{2})^{T}Q_{i,j}(\mathbf{u}_i+\mathbf{u}_j)
\end{split}
\end{equation}
where $Q_{i,j}\in \mathbb{R}^{d\times d}$ is the positive-definite symmetric matrix computed from the ellipsoid approximation of the LOS line segment between $\mathbf{x}_i,\mathbf{x}_j$. 
Thus it is straightforward that $h_{i,j,o}^\mathrm{los}(\mathbf{x},\mathbf{x}^\mathrm{obs})$ (a) is continuous differentiable, and (b) is of relative degree one. To verify the third condition (c), we substitute $\dot{h}_{i,j,o}^\mathrm{los}(\mathbf{x},\mathbf{x}^\mathrm{obs},\mathbf{u})=\frac{d}{dt}h_{i,j,o}^\mathrm{los}(\mathbf{x},\mathbf{x}^\mathrm{obs})$ from (\ref{d}) in $\dot{h}^\mathrm{los}_{i,j,o}(\mathbf{x}, \mathbf{x}^\mathrm{obs}, \mathbf{u})+\gamma h^\mathrm{los}_{i,j,o}(\mathbf{x}, \mathbf{x}^\mathrm{obs}) \geq 0$ from (\ref{eq:los_cbc_definition}), then we can obtain
\begin{align}\begin{aligned}
\label{provefeasibility}
(\mathbf{x}^\mathrm{obs}_{o}-\frac{\mathbf{x}_i+\mathbf{x}_j}{2})^{T}Q_{i,j}(\mathbf{u}_i+\mathbf{u}_j) \leq \gamma[(\mathbf{x}^\mathrm{obs}_{o}-\frac{\mathbf{x}_i+\mathbf{x}_j}{2})^{T}Q_{i,j}(\mathbf{x}^\mathrm{obs}_{o}-\frac{\mathbf{x}_i+\mathbf{x}_j}{2})-1]
\end{aligned}\end{align}
Note that without loss of generality, here the specific class-$\mathcal{K}$ function is chosen as $\kappa(h(\mathbf{x}))=\gamma h(\mathbf{x})$ with $\gamma > 0$ similar to \cite{wang2017safety}. Then in the following, we will prove there exists solutions of pairwise $\mathbf{u}_i,\mathbf{u}_j$ satisfying (\ref{provefeasibility}).
Since the considered pairwise robots $i,j$ are initially LOS connected (i.e. by definition $\forall (v_i,v_j)\in \mathcal{E}^\mathrm{slos}$ with $\mathcal{G}^\mathrm{slos}=(\mathcal{V},\mathcal{E}^\mathrm{slos})\subseteq \mathcal{G}^\mathrm{los}$), we have the right hand side of (\ref{provefeasibility}): $\gamma[(\mathbf{x}^\mathrm{obs}_{o}-\frac{\mathbf{x}_i+\mathbf{x}_j}{2})^{T}Q_{i,j}(\mathbf{x}^\mathrm{obs}_{o}-\frac{\mathbf{x}_i+\mathbf{x}_j}{2})-1]>0$, i.e. all the discretized obstacle points are not blocking the LOS between robots $i,j$. With that, it is impossible to find such a solution $y\geq 0\in\mathbb{R}$ so that $\{\gamma[(\mathbf{x}^\mathrm{obs}_{o}-\frac{\mathbf{x}_i+\mathbf{x}_j}{2})^{T}Q_{i,j}(\mathbf{x}^\mathrm{obs}_{o}-\frac{\mathbf{x}_i+\mathbf{x}_j}{2})-1]\}y<0$, implying that in our inequality system of (\ref{provefeasibility}) the second statement in Lemma~\ref{lem:farkas} is not true. To that end, the first statement in Lemma~\ref{lem:farkas} has to be true for our inequality system, proving that there exist solution of pairwise $\mathbf{u}_i,\mathbf{u}_j$ satisfying (\ref{provefeasibility}), i.e. the third condition (c) $\sup_{\mathbf{u}\in\mathcal{U}}\{\dot{h}^\mathrm{los}_{i,j,o}(\mathbf{x}, \mathbf{x}^\mathrm{obs}, \mathbf{u})+\gamma h^\mathrm{los}_{i,j,o}(\mathbf{x}, \mathbf{x}^\mathrm{obs})\} \geq 0$ proved to be true for our candidate CBF $h_{i,j,o}^\mathrm{los}(\mathbf{x},\mathbf{x}^\mathrm{obs})$. As a result, our candidate CBF $h_{i,j,o}^\mathrm{los}(\mathbf{x},\mathbf{x}^\mathrm{obs})$ in (\ref{los1}) is a valid CBF.

Secondly, we will show that the admissible control space defined by the LOS-CBC $\mathcal{B}^\mathrm{los}(\mathbf{x},\mathcal{C}_\mathrm{obs},\mathcal{G}^\mathrm{slos})$ in (\ref{eq:los_cbc_definition}) is always non-empty. Note that the corresponding control constraints consist of (i) $\mathcal{B}^c(\mathbf{x},\mathcal{G}^\mathrm{slos})$ in (\ref{eq8})-a set of connectivity constraints in terms of inter-robot limited communication distance constraints, and (ii) occlusion-free conditions in (\ref{provefeasibility}), for all pairwise robots $\forall (v_i,v_j)\in \mathcal{E}^\mathrm{slos}$ and discretized obstacles $\forall o\in\{1,\ldots,F\}$.

Hence, to ensure that the team of robots is LOS connected with a specific global and subgroup LOS connected graph $\mathcal{G}^\mathrm{slos}$, we can always formulate the linear constraints on the joint control space $\mathbf{u}\in \mathbb{R}^{dN}$ as $A^\mathrm{sys} \mathbf{u}\leq b^\mathrm{sys}$ for the team of robots, where,
\begin{equation}\label{app:eq:asys}
A^\mathrm{sys}= \begin{bmatrix}&0\quad ,...,&2(\mathbf{x}_i-\mathbf{x}_j)^T,& \ldots,  &-2(\mathbf{x}_i-\mathbf{x}_j)^T\quad&,...,\quad&0\\&0\quad,...,&(\mathbf{x}^\mathrm{obs}_{o}-\frac{\mathbf{x}_i+\mathbf{x}_j}{2})^{T}Q_{i,j},& \ldots, &(\mathbf{x}^\mathrm{obs}_{o}-\frac{\mathbf{x}_i+\mathbf{x}_j}{2})^{T}Q_{i,j}&\quad,...,\quad&0\\
&\vdots\quad,...,&\vdots,& \vdots, &\vdots&\quad,...,\quad&\vdots \end{bmatrix}
\end{equation}
\begin{equation}\label{app:eq:bsys}
b^\mathrm{sys}= \begin{bmatrix}\gamma(R^2_\mathrm{c}-||\mathbf{x}_i-\mathbf{x}_j||^{2})\\ \gamma[(\mathbf{x}^\mathrm{obs}_{o}-\frac{\mathbf{x}_i+\mathbf{x}_j}{2})^{T}Q_{i,j}(\mathbf{x}^\mathrm{obs}_{o}-\frac{\mathbf{x}_i+\mathbf{x}_j}{2})-1]\\
\vdots
\end{bmatrix}
\end{equation}
where $A^\mathrm{sys}\in \mathbb{R}^{z\times dN}$, $b^\mathrm{sys}\in\mathbb{R}^{z}$ and $z$ is the total number of the control constraints due to LOS-CBC for the entire system. Similarly, we have that $b^\mathrm{sys}\in\mathbb{R}^{z}$ in (\ref{app:eq:bsys}) is always larger than 0 given the initial LOS connectivity condition for $\forall (v_i,v_j)\in \mathcal{E}^\mathrm{slos}$
% , i.e. all the team of robots with the specific $\mathcal{G}^\mathrm{slos}$ graph satisfies the LOS connectivity condition initially
. With that, it is impossible to find a solution $y \geq 0 \in \mathbb{R}^{z}$ so that $y^\top b^\mathrm{sys}<0$, indicating that in our entire inequality system $A^\mathrm{sys} \mathbf{u}\leq b^\mathrm{sys}$, the second statement in Lemma~\ref{lem:farkas} is not true. With this, the first statement in Lemma~\ref{lem:farkas} must be true for our $A^\mathrm{sys} \mathbf{u}\leq b^\mathrm{sys}$, proving that there exist solution of joint control $\mathbf{u}$ satisfying~(\ref{eq:los_cbc_definition}). To that end, we can prove the admissible control space constrained by~(\ref{eq:los_cbc_definition}) is always non-empty. Thus, it concludes the proof.

\end{proof}

\subsection{Proof of Lemma~\ref{loscbcdefinition}}\label{app:sec:los-cbc}
\begin{customlemma}{2}
\textbf{Line-of-Sight Connectivity Barrier Certificates(LOS-CBC):}
Given a LOS communication spanning graph $\mathcal{G}^\mathrm{slos}=(\mathcal{V},\mathcal{E}^\mathrm{slos})\subseteq \mathcal{G}^\mathrm{los}$ and a desired set $\mathcal{H}^\mathrm{los}(\mathcal{G}^\mathrm{slos})$ in (\ref{hlos}) with $h^\mathrm{los}_{i,j,o}$ from (\ref{los1}), the Line-of-Sight connectivity barrier certificates (LOS-CBC) as the admissible control space $ \mathcal{B}^\mathrm{los}(\mathbf{x},\mathcal{C}_\mathrm{obs},\mathcal{G}^\mathrm{slos})$ defined below renders $\mathcal{H}^\mathrm{los}(\mathcal{G}^\mathrm{slos})$ forward invariant (i.e keeping joint robot state staying in $\mathcal{H}^\mathrm{los}(\mathcal{G}^\mathrm{slos})$):%
\vspace{-5pt}
{\begin{align}\tag{17}
\begin{aligned}
&\mathcal{B}^\mathrm{los}(\mathbf{x},\mathcal{C}_\mathrm{obs},\mathcal{G}^\mathrm{slos})= \mathcal{B}^c(\mathbf{x},\mathcal{G}^\mathrm{slos})\bigcap \{\mathbf{u}\in \mathbb{R}^{dN}:\\
&\;\dot{h}^\mathrm{los}_{i,j,o}(\mathbf{x}, \mathbf{x}^\mathrm{obs}, \mathbf{u})+\gamma h^\mathrm{los}_{i,j,o}(\mathbf{x},\mathbf{x}^\mathrm{obs})\geq 0,\forall (v_i,v_j)\in \mathcal{E}^\mathrm{slos},\forall o\} 
%\in \partial{\mathcal{C}_\mathrm{obs}}\} 
\end{aligned}
\end{align}}%
where $\dot{h}_{i,j,o}^\mathrm{los}(\mathbf{x},\mathbf{x}^\mathrm{obs}, \mathbf{u}) = -(\mathbf{x}^\mathrm{obs}_{o}-\frac{\mathbf{x}_i+\mathbf{x}_j}{2})^{T}Q_{i,j}(\mathbf{u}_{i}+\mathbf{u}_{j}).$ 
\end{customlemma}%

\begin{proof}
    
In order to maintain the team of robots to stay LOS connected. It is necessary to keep required edges in $\mathcal{G}^\mathrm{slos}$ LOS connected. Then the desired LOS connectivity set for the team of robots can be defined as:

{\begin{align}
    \mathcal{H}^\mathrm{los}(\mathcal{G}^\mathrm{slos})= \Big(\bigcap_{\{v_i,v_j \in \mathcal{V}:(v_i,v_j)\in\mathcal{E}^\mathrm{slos}\}} \mathcal{H}^\mathrm{los}_{i,j} \Big)\bigcap \mathcal{H}^\mathrm{c}(\mathcal{G}^\mathrm{slos})\tag{10}
\end{align}}% 

The $\mathcal{H}^\mathrm{los}(\mathcal{G}^\mathrm{slos})$ is the intersection set of (i) the desired sets for the team of robots satisfy the connectivity distance constraints and (2) the desired sets for the team of robots satisfy the occlusion-free condition. To guarantee the intersection set forward invariant (i.e. keep the team of robots line-of-sight connected all the time), we proposed the LOS-CBC in~(\ref{eq:los_cbc_definition}). Note that, in Lemma~\ref{valid}, we have already proved that the constrained space by~(\ref{eq:los_cbc_definition}) is always not empty. 
%If the controller $\mathbf{u}$ satisfy the constraints in~(\ref{eq:los_cbc_definition}), then 

The proposed LOS-CBC aims to render an intersection set forward invariant. Adopting the CBF related method to keep an intersection set forward invariant has been studied in \cite{li2018formally}. Considering the property of the control barrier functions in Lemma~\ref{lem:cbf}, the admissible control space which can satisfy the LOS-CBC can keep each set in(~\ref{hlos}) forward invariant (keeping the system state $\mathbf{x}$ staying in each set in(~\ref{hlos}) over time).  Hence, by enforcing the control input in the intersection admissible control space constrained by our LOS-CBC, it can guarantee the intersection set $\mathcal{H}^\mathrm{los}(\mathcal{G}^\mathrm{slos})$ in~(\ref{hlos}) forward invariant. Thus, we conclude the proof. 

%Hence, the proposed LOS-CBC can guarantee that the system state $\mathbf{x}$ stays in the following set:
%\begin{equation}\label{app:eq:hunion}
 %   \mathcal{H}^\mathrm{sys}_{\bigcap} = \{\mathbf{x} \in \mathbb{R}^{d}: \prod_{\forall (v_i,v_j)\in\mathcal{E}^\mathrm{slos}}{h^\mathrm{c}_{i,j}}(\mathbf{x})\prod_{\forall (v_i,v_j)\in\mathcal{E}^\mathrm{slos},\forall o}{h^\mathrm{los}_{i,j,o}(\mathbf{x})} \geq 0 \}
%\end{equation}

%In Lemma~\ref{lemma:hintersection}, it can be extended to the form of multiplication of multiple valid control barrier functions \cite{wang2016multi}. With the extension format of Lemma~\ref{lemma:hintersection}, it implies $\mathcal{H}^\mathrm{sys}_{\bigcap} =\mathcal{H}^\mathrm{los}(\mathcal{G}^\mathrm{slos})$. Hence, by enforcing the control input in the admissible control space constrained by our LOS-CBC, it can guarantee the intersection set $\mathcal{H}^\mathrm{los}(\mathcal{G}^\mathrm{slos})$ in~(\ref{hlos}) forward invariant. Thus, we conclude the proof. 

%
%hen with Lemma~\ref{lemma:hintersection} and Theorem~\ref{theorem:intersection}, it is clearly to show that our LOS-CBC in~(\ref{eq:los_cbc_definition}) could ensure the intersection set $\mathcal{H}^\mathrm{los}(\mathcal{G}^\mathrm{slos})$ in~(\ref{hlos}) forward invariant. In another word, if the team of the robots are initially LOS connected with a specific graph $\mathcal{G}^\mathrm{slos}$, then our proposed LOS-CBC could ensure the team of the robots stay LOS connected. 

\end{proof}

\subsection{Feasibility Discussion of the QP Problem}\label{app:sec:feasible_qp}
In this paper, the composition of different valid control barrier functions can be defined as:

% \begin{footnotesize}
\begin{equation}
    h_{i,j,o}^\mathrm{sys} = h_{i,j,o}^\mathrm{los} \land h_{i,j}^\mathrm{c}\land h_{i,j}^\mathrm{s}\land h_{i,o}^\mathrm{obs}= \min\{{\min\{{h_{i,j,o}^\mathrm{los},h_{i,j}^\mathrm{c}}\},\min\{{h_{i,j}^\mathrm{s},h_{i,o}^\mathrm{obs}}\}}\}\tag{23}
\end{equation}
% \end{footnotesize}
Note the corresponding control constraints consist of (i) The LOS-CBC $\mathcal{B}^\mathrm{los}(\mathbf{x},\mathcal{C}_\mathrm{obs},\mathcal{G}^\mathrm{slos})$ in (\ref{eq:los_cbc_definition}) - a set of pair-wise robot LOS connectivity constraints, (ii) the $\mathcal{B}^\mathrm{s}(\mathbf{x})$ in~(\ref{eq7})- a set of inter-robot safety constraints, and (iii) the $\mathcal{B}^\mathrm{obs}(\mathbf{x},\mathbf{x}^\mathrm{obs})$ in~(\ref{bobs}) - a set of robot-discretized obstacles safety constraints. The composition of CBFs is studied in \cite{capelli2020connectivity,egerstedt2018robot,glotfelter2017nonsmooth}. 
% \yy{version 1}
% We can stack all the constraints as $A^\mathrm{sys}\mathbf{u} \leq b^\mathrm{sys}$, where $A^\mathrm{sys}\in \mathbb{R}^{m\times qN}$, $b^\mathrm{sys} \in \mathbb{R}^{m}$ and $m$ is the total number of constraints in our system. Since the team of robots satisfy the initial safety and LOS connectivity condition (i.e. $b^\mathrm{sys} >0$), it is impossible to find such a solution $y>=0$ so that $b^\mathrm{sys}y<0$ implying that in our inequality system of $A^\mathrm{sys}\mathbf{u} \leq b^\mathrm{sys}$, the second statement in Lemma~\ref{lem:farkas} is not true. Hence the first statement of Lemma~\ref{lem:farkas} is true for our inequality system, proving that their exist solution $\mathbf{u}$ satisfying $A^\mathrm{sys}\mathbf{u} \leq b^\mathrm{sys}$. \\
%\begin{remark}(Feasibility for bounded input constraints)
%Consider the bounded input constraints, (i.e. $\mathcal{U} := \{\mathbf{u}_i \in \mathbb{R}^{d}:||\mathbf{u}_{i}|| \leq %\mathbf{u}_{max}\}$), the feasible set constrained by $A^\mathrm{sys}\mathbf{u} \leq b^\mathrm{sys}$ could be empty. In %\cite{lyu2021probabilistic}, we provided an optimization solution for parameter $\gamma$ over time to guarantee the feasible set %constrained by the CBF constraints and the input $\mathcal{U}$ constraints is always not empty. 
%\end{remark}
Those three control constraints from different valid control barrier functions indicate that $\mathbf{u} =0$ is their non-empty interior. Considering the unbounded input, it means that the feasible set constrained by~(\ref{eq:rawconst}) is always not empty.
In presence of bounded input constraints, 
% \yy{(
i.e. $\mathcal{U}_i := \{\mathbf{u}_i \in \mathbb{R}^{d}: u_\mathrm{min}\leq||\mathbf{u}_{i}|| \leq u_\mathrm{max}\}$,
% )}
the feasible set constrained by~(\ref{eq:rawconst}) could be empty. 
% \yy{
For example, due to the physical limitations of the self-driving vehicle's braking system, it is not possible for the vehicle to instantaneously decelerate to a complete stop.
% }
In \cite{lyu2021probabilistic}, we provided an optimization solution for varying parameter $\gamma$ over time to adaptively adjust the admissible control space by the CBF constraints, so that the feasibility is maximized.
% the input $\mathcal{U}$ constraints is always not empty.
Moreover, the authors in \cite{xiao2022sufficient} provided a novel method to find sufficient conditions, which 
% They derived sufficient conditions which 
are captured by a single constraint and enforced by an additional CBF, to guarantee the feasibility of original CBF-based QPs.
% , to guarantee the feasibility of the original CBF-based QPs. 
Note that the additional CBF will always be compatible with the existing constraints, implying that it cannot make the previous feasible set of constraints infeasible. Readers are referred to \cite{xiao2022sufficient} for further details.
% With that, we can prove the feasibility of our proposed QP problem. 

\subsection{Proof of Theorem~\ref{theorem:mlccst}}\label{app:sec:mlccst_theorem}
\begin{customthm}{4}
Given the redefined Line-of-Sight Spanning Tree (LOS-ST) $\mathcal{T}_w^\mathrm{los'}=(\mathcal{V},\mathcal{E}^{T},\mathcal{W}^{T'})$ in Definition \ref{def:lccst} and denote minimum weighted LCCST as $\bar{\mathcal{T}}_w^\mathrm{los'}=\argmin_{\{\mathcal{T}_w^\mathrm{los'}\}} \sum_{(v_i,v_j)\in \mathcal{E}^{T}}\{-w'_{i,j}\}$, then we have: $\mathcal{G}^\mathrm{slos*}=\bar{\mathcal{T}}_w^\mathrm{los'}$ in (\ref{eq:expobj}). Namely, the Minimum Spanning Tree $\bar{\mathcal{T}}_w^\mathrm{los'}$ of $\mathcal{G}^\mathrm{los'}$ is the optimal solution of $\mathcal{G}^\mathrm{slos*}$ in (\ref{eq:expobj}) and we call the graph $\bar{\mathcal{T}}_w^\mathrm{los'}$ asUncertainty-Aware Line-of-Sight Minimum Spanning Tree (LOS-MST) of the original LOS communication graph $\mathcal{G}^\mathrm{los}$.
\end{customthm}
From Definition~\ref{def:lccst}, we can ensure the weight for each edge $(v_i,v_j), (v_{i'},v_{j'})\in\mathcal{E}$ in graph $\mathcal{G}'$ has the following relation:
\begin{align}
\label{eq:relation}
-\lambda \cdot w^\mathrm{d+los}_{i,j}< -w^\mathrm{d+los}_{i',j'} \ll -\epsilon < -\lambda\cdot\epsilon
\end{align}
where $\lambda \gg 1, \epsilon \ll 0,\forall v_i,v_i',v_j,v_j'\in \mathcal{V}$ has been defined after (\ref{WLOSdefine}) and (\ref{eq:neww}). 

Then we briefly review some useful definitions in  graph theory \cite{gallager1983distributed}:
1) fragment: a subtree of Minimum Spanning Tree; 2) outgoing edge: an edge of a fragment if one adjacent node is in the fragment and the other is not; 3) minimum-weight outgoing edge (MWOE):an edge with minimum weight among all outgoing edge of a fragment.  
%\begin{itemize}
%\item \textbf{outgoing edge}: an edge of a fragment if one adjacent node is in the fragment and the other is not;
%\item \textbf{minimum-weight outgoing edge (MWOE)}: an edge with minimum weight among all %outgoing edge of a fragment.  
%\end{itemize}

In \cite{gallager1983distributed} and \cite{peleg2000distributed}, it has been proved that connecting the minimum-weight outgoing edge (MWOE) and its adjacent node in a different fragment yields another fragment in MST. Note that MST is unique for a graph with unique edge weights. Hence they construct the MST in the following process: 1) Each node starts as a fragment by itself, and 2) each fragment iteratively connects with MWOE fragment.

\begin{proof}
Firstly, we will prove that our proposed MLCCST will guarantee that the LOS edges \textbf{between} sub-groups will be LOS connected only when LOS edges \textbf{within} each sub-groups are LOS connected.

We prove this by contradiction. Suppose the node $v_i$ from sub-group graph $\mathcal{G}_i^\mathrm{los'}$ LOS connects with node $v_j$ first, which belongs to sub-group graph $\mathcal{G}_j^\mathrm{los'}$, $i \neq j$ to form the optimal spanning tree $\bar{\mathcal{T}}_w^\mathrm{los'}$. From the MST construction process we know that, at each iteration, the edge added is the minimum-weight outgoing edge of the connecting fragment. In this case, the weight $\mathcal{W}'_{i,j}$ of the edge between $v_i$ and $v_j$ is the minimum of all outgoing edges of $v_i$.
Let $v_{k} \in \mathcal{G}_i^\mathrm{los'}$ where there exists an outgoing edge between $v_i$ and  $v_{k}$, then we know that the weight $\mathcal{W}'_{i,j} < \mathcal{W}'_{i,k}$. This contradicts with the property of $\mathcal{G}^\mathrm{los'}$ in Equation (\ref{eq:relation}). 

With that by Definition~\ref{def:lccst}, the MST of graph $\mathcal{G}_i^\mathrm{los'}$ within subgroup $\mathcal{S}_i$ which is LOS connected is optimal with minimum total weight, which means :
\begin{equation}
\begin{split}
     \bar{\mathcal{T}}_w^\mathrm{los'}(i)&=\argmin_{\mathcal{T}_w^\mathrm{los'}(i)\in \mathcal{T}(i)} \sum_{(v_i,v_j)\in \mathcal{E}^{T(i)}}-w'_{i,j}\\
     &=\argmin_{\mathcal{T}_w^\mathrm{los'}(i)\in \mathcal{T}(i)} \lambda\cdot\sum_{(v_i,v_j)\in \mathcal{E}^{T(i)}}-w^\mathrm{d+los}_{i,j}\\
     &=\argmin_{\mathcal{T}_w^\mathrm{los'}(i)\in \mathcal{T}(i)} \sum_{(v_i,v_j)\in \mathcal{E}^{T(i)}}-w^\mathrm{d+los}_{i,j}
\end{split}
\end{equation}
The equality holds since $\lambda > 0$. Then we consider $v_i$ and $v_j$ in different subgroups, i.e. $\mathcal{S}(v_i) \neq \mathcal{S}(v_j)$, while $(v_i, v_j)$ is the edge in spanning tree edges $\mathcal{E}^{T(i)}$ LOS connecting two subgroups. Then for the next step, connecting the minimum-weighted outgoing edge between different sub-groups yields:
\begin{equation} 
\begin{split}
    \bar{\mathcal{T}}_w^\mathrm{los'}
     &=\argmin_{\mathcal{T}_w^\mathrm{los'}\in \mathcal{T}} \sum_{(v_i,v_j)\in \mathcal{E}^{T(i)}}-w'_{i,j}, \quad \mathcal{S}(v_i) \neq \mathcal{S}(v_j)\\
    &=\argmin_{\mathcal{T}_w^\mathrm{los'}\in \mathcal{T}} \sum_{(v_i,v_j)\in \mathcal{E}^{T}}-w^\mathrm{d+los}_{i,j}
 \end{split} 
 \end{equation}
%\begin{equation}
%\begin{split}
 %    \bar{\mathcal{T}}_w^\mathrm{los'}
  %   &=\argmin_{\mathcal{T}_w^\mathrm{los'}\in \mathcal{T}} \sum_{(v_i,v_j)\in \mathcal{E}^{T(i)}}-w'_{i,j}, \quad \mathcal{S}(v_i) %\neq \mathcal{S}(v_j)\\
   %  &=\argmin_{\mathcal{T}_w^\mathrm{los'}\in \mathcal{T}} \sum_{(v_i,v_j)\in \mathcal{E}^{T}}-w^\mathrm{d+los}_{i,j}
%\end{split} 
%\end{equation}
With the same form as in (\ref{eq:expobj}), this concludes the proof.\\
% In this way, the edges which are occluded by the obstacles will be selected only when LOS connectivity edges are selected. 
\end{proof}

\subsection{Proof of Proposition~\ref{guarantee}}\label{app:LOS_overtime}
\begin{figure}[!htbp]
    \centering
    \includegraphics[width =0.7\textwidth]{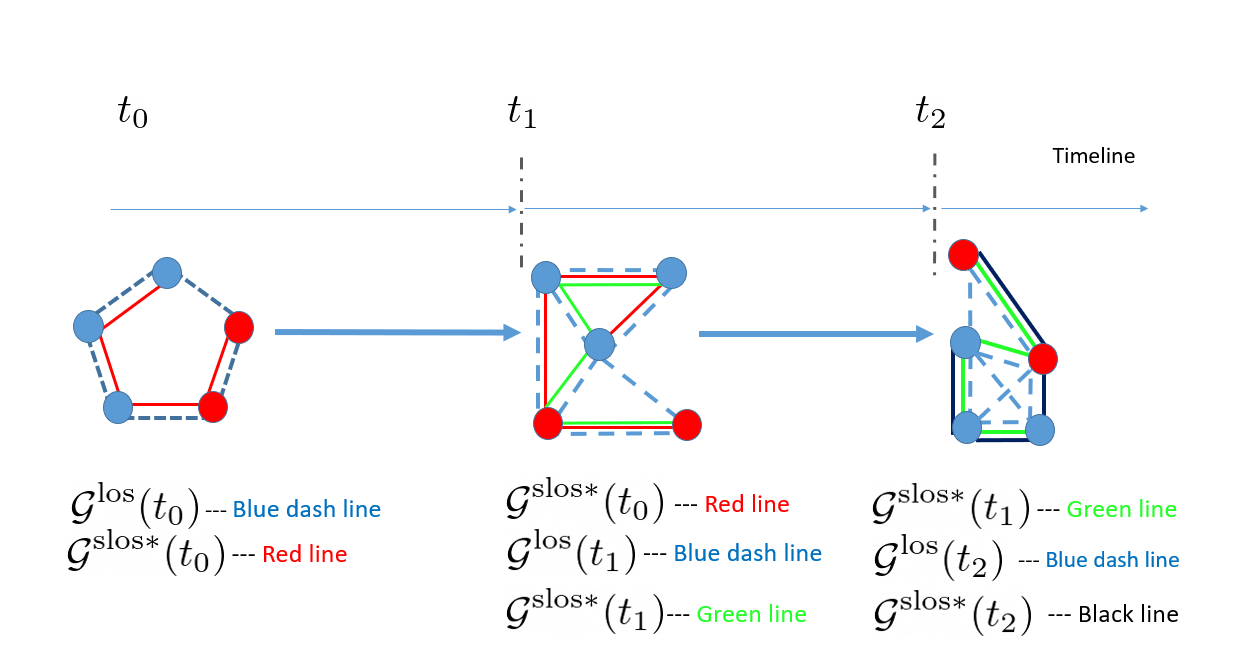}
    \caption{Sequence of the proposed MLCCST algorithm.}
    \label{fig:app:graph}
\end{figure}

\begin{customprop}{6}
Assume $ \mathcal{G}^\mathrm{los}$ is initially both global and subgroup LOS connected. By 
% assigning the weights in (\ref{WLOSdefine}) and 
following the process 
% of 
% constructing MLCCST $\bar{\mathcal{T}}_w^\mathrm{los'}\subseteq \mathcal{G}^\mathrm{los'}$ 
in Algorithm~\ref{alg:Dynamic} at each time step, it is guaranteed that the resulting communication graph $\mathcal{G}^\mathrm{los}$ in the next time step
% 1) MLCCST $\bar{\mathcal{T}}_w^\mathrm{los'}$ is the spanning sub-graph of the LOS connectivity graph (i.e.$\bar{\mathcal{T}}_w^\mathrm{los'} \subseteq \mathcal{G}^\mathrm{los}$), and 2) the MLCCST $\bar{\mathcal{T}}_w^\mathrm{los'}$ 
is always global and subgroup LOS connected.
\end{customprop}
\begin{remark}
The LOS communication graph in our paper is time-dependent, which means the LOS communication graph can be changed over time (e.g. $\mathcal{G}^\mathrm{los}=\mathcal{G}^\mathrm{los}(t)$). In the main text, we omit dependence on time, for ease of notation. To prove Proposition 6, we give the explicit form of the LOS commendation graph with respect to the time $t$. 
\end{remark}
\begin{proof}
If the team of robots is both global and subgroup LOS connected initially, there exists a solution to form an MLCCST $\bar{\mathcal{T}}_w^\mathrm{los'}(t = t_0)$ where all edges therein are both global and subgroup LOS connected (i.e. $\bar{\mathcal{T}}_w^\mathrm{los'}(t = t_0)\subseteq \mathcal{G}^\mathrm{los}(t=t_0)$). 
%and the MLCCST $\bar{\mathcal{T}}_w^\mathrm{los'}(t_0)$ is global and subgroup LOS connected at the first time step.
It is shown in Algorithm~\ref{alg:Dynamic}, after constructing the MLCCST $\bar{\mathcal{T}}_w^\mathrm{los'}(t) = \bar{\mathcal{T}}_w^\mathrm{los'}(t = t_0)$ within $t\in[t_0,t_0+\tau]$, the robots are constrained by the LOS-CBC with the current MLCCST $\bar{\mathcal{T}}_w^\mathrm{los'}(t= t_0)$. The LOS-CBC is the step-wise optimization method that depends on the current graph $\bar{\mathcal{T}}_w^\mathrm{los'}(t= t_0)$. Our controller $\mathbf{u}(\mathbf{x})$ depends on each robot state $\mathbf{x}$ and if the Lipschitz continuous controller $\mathbf{u}(\mathbf{x}(t)) \in \mathcal{B}^\mathrm{los}(\mathbf{x},\mathcal{C}_\mathrm{obs},\bar{\mathcal{T}}_w^\mathrm{los'}(t= t_0))$ for all $t\in [t_0,t_0 +\tau]$, then it guarantees  $\mathbf{x}(t) \in \mathcal{H}^\mathrm{los}(\mathcal{G}^\mathrm{slos}(t))$ within $t\in [t_0,t_0+\tau]$ (see lemma~\ref{loscbcdefinition}). In another word, this guarantees that the $\bar{\mathcal{T}}_w^\mathrm{los'}(t= t_0)$ is both global and subgroup LOS connected within $t\in [t_0,t_0+\tau]$. Then at the next time-step $t_1 = t_0 + \tau$, it is guaranteed that $\bar{\mathcal{T}}_w^\mathrm{los'}(t=t_0) \subseteq \mathcal{G}^\mathrm{los}(t=t_1)$. As shown in Figure~\ref{fig:app:graph}, the MLCCST algorithm can guarantee that the preserved optimal spanning tree $\bar{\mathcal{T}}_w^\mathrm{los'}(t) = \mathcal{G}^\mathrm{slos*}(t)$ is always the subgraph of the resulting communication graph $\mathcal{G}^\mathrm{los}(t)$. In summary, since the step-wise optimal spanning tree $\bar{\mathcal{T}}_w^\mathrm{los'}(t)$ is global and subgroup LOS connected shown in (~\ref{eq:rawglobal} and \ref{eq:rawconn}), then the parent graph $\mathcal{G}^\mathrm{los}(t)$ is always global and subgroup LOS connected. Thus we conclude the proof.
%\wl{We should add something like: since the step-wise optimal spanning tree T is global and subgroup LOS connected shown in (the equation of the QP problem), then its supergraph Glos(t) is always global and subgroup LOS connected. Thus we conclude the proof.}
%which indicates that there will always exist at least one solution to form the MLCCST $\bar{\mathcal{T}}_w^\mathrm{los'}(t_1)\in \mathcal{G}^\mathrm{los}(t_1)$ with all its edges global and subgroup LOS connected for the next time step. 
\end{proof}
\end{document}